%% file: acl_latex.tex
\documentclass[11pt]{article}
\usepackage{amsfonts}
\usepackage[preprint]{acl}

\usepackage{times}
\usepackage{latexsym}

\usepackage[T1]{fontenc}
\usepackage{booktabs}
\usepackage{multirow}
\usepackage{amsmath}
\usepackage{xcolor}
\usepackage{algorithm}
\usepackage[most]{tcolorbox}
\usepackage[utf8]{inputenc}

\usepackage{microtype}

\usepackage{inconsolata}

\usepackage{graphicx}

\usepackage{amsmath}
\usepackage{xspace}
\usepackage{booktabs}
\usepackage{colortbl}
\usepackage{multirow}
\usepackage{xcolor}
\usepackage[most]{tcolorbox}
\usepackage{xcolor}
\usepackage{tabularx}
\usepackage{listings}
\usepackage{enumitem}
\usepackage{fontawesome5}

\newcommand{\method}{Parametric Skill Transfer\xspace}
\newcommand{\methodFull}{Parametric Skill Transfer (PaST)\xspace}
\newcommand{\methodShort}{PaST\xspace}
\newcommand{\gain}[1]{\rlap{~\scriptsize{\color{green!40!black}(+#1)}}}

\usepackage[most]{tcolorbox}
\usepackage{listings}
\usepackage{tabularx}

\lstdefinestyle{trajectoryStyle}{
    basicstyle=\ttfamily\small,
    breaklines=true,
    breakindent=0pt,
    breakatwhitespace=true,
    columns=fullflexible,
    keepspaces=true,
    literate={_}{\_}{1} {\{}{\textbraceleft}{1} {\}}{\textbraceright}{1},
    moredelim=[is][\bfseries]{|}{|},
}

\newtcolorbox{mybox}[2][]{
    enhanced,
    colback=white,
    colframe=#2!70!black,
    fonttitle=\bfseries,
    title=#1,
    arc=1mm,
    boxrule=0.8pt,
    left=2mm, right=2mm, top=2mm, bottom=2mm,
    width=\textwidth
}

\title{Knowledge is Not Enough: Injecting RL Skills for Continual Adaptation}

\usepackage{marvosym}
\author{
Pingzhi Tang\textsuperscript{$*$1,2}, 
Yiding Wang\textsuperscript{$*$1,2}, 
Muhan Zhang\textsuperscript{1}\\
\textsuperscript{1}Institute for Artificial Intelligence, Peking University \quad 
\textsuperscript{2}Yuanpei College, Peking University \\
\textsuperscript{*}Equal contribution \quad 
\Letter\ Correspondence to \href{mailto:muhan@pku.edu.cn}{muhan@pku.edu.cn} \\
\faGithub\,\href{https://github.com/MuLabPKU/PaST}{\texttt{MuLabPKU/PaST}}
}

\begin{document}
\maketitle

\input{src/abstract}
\input{src/introduction}
\input{src/related_work}

\input{src/motivation}
\input{src/method}
\input{src/experiments}

\input{src/conclusion}

\newpage
\input{src/limitation}
\input{src/acknowledgement}

{\sloppy\bibliography{custom}\par}

\input{src/appendix/main}

\end{document}

%% file: src/abstract.tex
\begin{abstract}
Large Language Models (LLMs) face the ``knowledge cutoff'' challenge, where their frozen parametric memory prevents direct internalization of new information. While Supervised Fine-Tuning (SFT) is commonly used to update model knowledge, it often updates factual content without reliably improving the model’s ability to use the newly incorporated information for question answering or decision-making. Reinforcement Learning (RL) is essential for acquiring reasoning skills; however, its high computational cost makes it impractical for efficient online adaptation. We empirically observe that the parameter updates induced by SFT and RL are nearly orthogonal. Based on this observation, we propose \textbf{Parametric Skill Transfer (PaST)}, a framework that supports modular skill transfer for efficient and effective knowledge adaptation.
By extracting a domain-agnostic \textbf{Skill Vector} from a source domain, we can linearly inject knowledge manipulation skills into a target model after it has undergone lightweight SFT on new data.
Experiments on knowledge-incorporation QA (SQuAD, LooGLE) and agentic tool-use benchmarks (ToolBench) demonstrate the effectiveness of our method.
On SQuAD, PaST outperforms the state-of-the-art self-editing SFT baseline by up to 9.9 points. PaST further scales to long-context QA on LooGLE with an 8.0-point absolute accuracy gain, and improves zero-shot ToolBench success rates by +10.3 points on average with consistent gains across tool categories, indicating strong scalability and cross-domain transferability of the Skill Vector.
\end{abstract}

%% file: src/introduction.tex
\section{Introduction}
\label{sec:intro}

\begin{figure*}[ht]
\centering
\includegraphics[width=\textwidth]{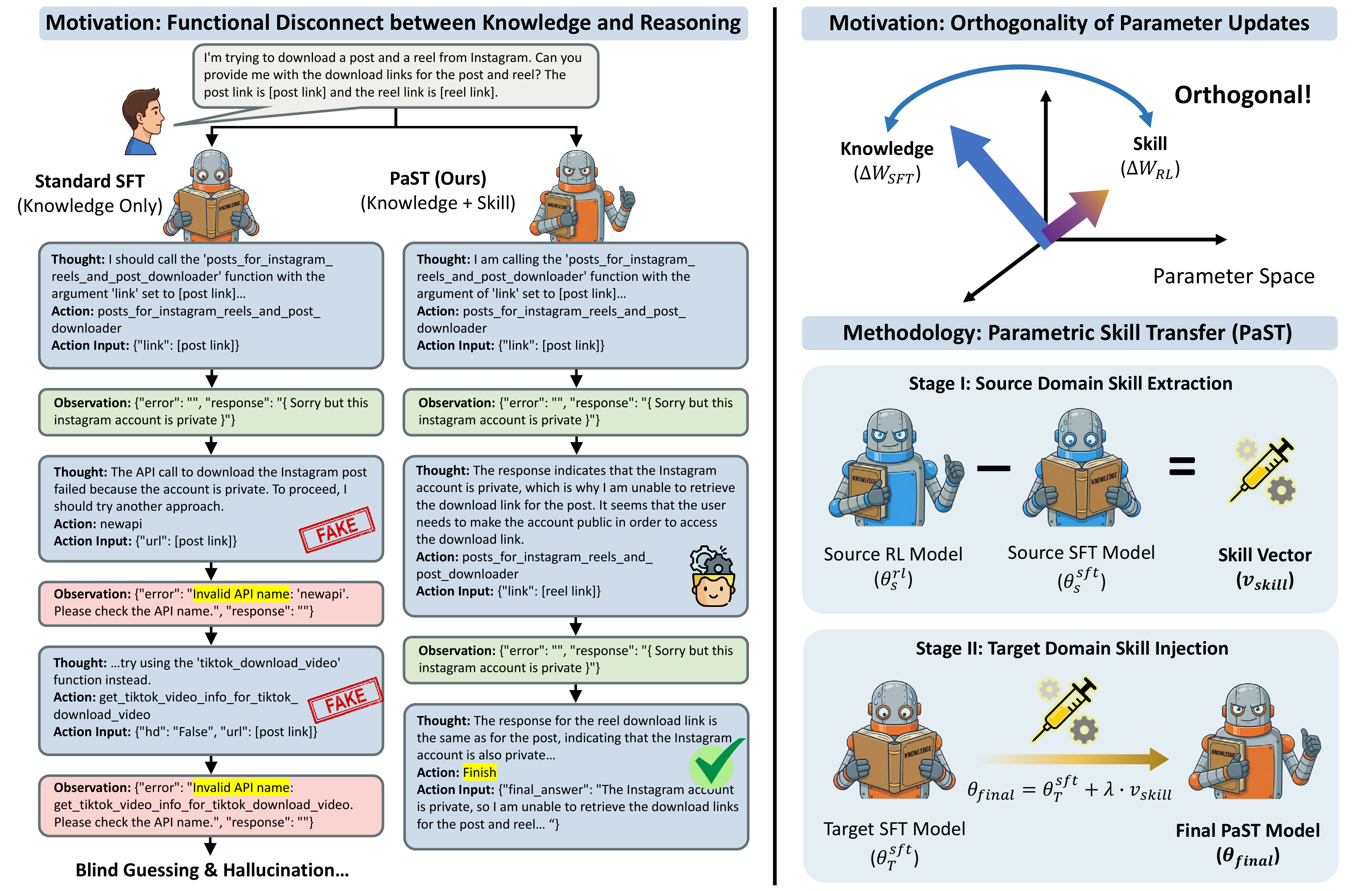}
\caption{\textbf{Overview of Parametric Skill Transfer (PaST).}
The motivation (left) illustrates how standard SFT fails to handle environmental errors, leading to hallucinations, while PaST enables robust execution by incorporating reasoning skills.
Our approach is based on the empirical finding (top right) that parameter updates for knowledge ($\Delta W_{SFT}$) and skills ($\Delta W_{RL}$) are nearly orthogonal and reside in disentangled subspaces.
PaST first extracts a domain-agnostic skill vector $v_{skill} = \theta_{S}^{rl} - \theta_{S}^{sft}$ from a source domain and then linearly injects it into a target model via $\theta_{final} = \theta_{T}^{sft} + \lambda \cdot v_{skill}$, enabling efficient and effective knowledge adaptation without requiring expensive reinforcement learning in the target domain.}
\label{fig:hero}
\end{figure*}

Large Language Models (LLMs) \cite{vaswani2017attention, brown2020language} have demonstrated remarkable capabilities in \textbf{static} benchmarks, yet their utilization in real-world scenarios is constrained by the ``knowledge cutoff'' problem \citep{cheng2024dated}---the inherent limitation that their parametric memory remains \textbf{frozen} after pre-training, preventing them from natively internalizing new information or tools on the fly~\citep{ouyang2022training, touvron2023llama}. 
Retrieval-Augmented Generation (RAG)~\citep{lewis2020retrieval} attempts to mitigate this by injecting external context at inference time; however, it often struggles with long-range dependency modeling over large corpora and incurs substantial inference-time overhead due to repeated processing of retrieved contexts\citep{shao2023enhancing}.
Consequently, recent research has shifted towards \textit{Parametric Knowledge Updating}, aiming to efficiently internalize new information directly into model weights. Techniques such as Knowledge Editing \citep{meng2022locating, yao2023editing, mao2025lift} and Test-Time Training (TTT) \citep{liu2021ttt++, osowiechi2023tttflow, gandelsman2022test, hong2023mecta} have emerged as promising directions, attempting to keep the model's parametric memory synchronized with the evolving world.

However, a critical limitation of existing adaptation paradigms is the functional disconnect between \textit{knowledge} and \textit{skills}.
Prevailing methods largely rely on Supervised Fine-Tuning (SFT) to inject new domain knowledge. 
Recent work \citep{chu2025sft} highlighted a fundamental distinction in optimization dynamics: SFT \textbf{memorizes}, RL \textbf{generalizes}.
Supported by our experiments, SFT tends to induce surface-level memorization of the training distribution, without explicitly teaching the model how to reason over the acquired knowledge in downstream tasks. While Reinforcement Learning (RL) is essential for acquiring robust reasoning and execution skills \citep{guo2025deepseek}, it remains a bottleneck for efficient online adaptation to novel scenarios. The high cost of collecting interaction data and the computational burden of on-policy exploration make it infeasible to perform RL for every new environment the model encounters.

To bridge this gap, we propose \textbf{\methodFull}, a modular framework that injects RL-optimized reasoning capabilities into models adapted to new knowledge, without explicitly performing RL on the new knowledge. 
Our approach is driven by the empirical observation that the parameter updates induced by SFT and RL occupy nearly orthogonal spaces.
Therefore, we hypothesize that the SFT and RL updates are natively decoupled, where the RL-learned skills need not be bound to specific SFT knowledge but can be transferred to new domains.
Subsequently, we introduce a mechanism to extract the \textbf{domain-agnostic} \textit{skill vector} by subtracting the parameters of an SFT-anchored model from its RL-refined counterpart in a source domain. This vector, which captures the ``gradient'' direction of reasoning improvement, can then be linearly added to a target model during the adaptation phase, immediately after it has undergone lightweight SFT on new target data, which avoids the expensive RL process in the new domain. Figure~\ref{fig:hero} illustrates the idea.

We empirically evaluate \methodShort across two primary capability domains: Knowledge Incorporation (covering both short- and long-context scenarios via SQuAD \citep{rajpurkar2016squad} and LooGLE \citep{li2024loogle}) and Closed-Book Tool Use (via ToolBench \citep{qin2023toolllm, guo2024stabletoolbench}). Extensive experiments demonstrate the efficacy of our framework.
First, on the SQuAD knowledge incorporation task, \methodShort achieves 56.9\% accuracy, surpassing the state-of-the-art self-adapting baseline SEAL (47.0\%, \citet{zweiger2025self}) by a substantial margin.
Second, on the LooGLE long-context benchmark, we demonstrate that our approach scales to massive documentation (over 24k tokens), enabling more precise information retrieval from parametric memory than standard SFT.
Finally, in ToolBench cross-domain evaluation, \methodShort enables zero-shot transfer of tool-use skills to RL-unseen categories, successfully activating execution capabilities in target domains.

Our contributions are summarized as follows:
\begin{itemize}[leftmargin=*, itemsep=0pt, parsep=0pt, topsep=0pt, partopsep=0pt]
    \item We identify the \textit{Reasoning-Knowledge Disconnect} in knowledge adaptation, highlighting the insufficiency of SFT for transferring procedural logic to new domains.
    \item We provide empirical evidence that parameter updates induced by skill learning (via RL) and knowledge acquisition (via SFT) are nearly orthogonal and reside in disentangled subspaces of the parameter landscape, enabling separate optimization and linear composition.
    \item We propose \methodFull, a novel method that utilizes task vector arithmetic to transfer RL-learned skills from a source domain to a target domain, bypassing the need for test-time RL.
    \item We empirically demonstrate the effectiveness of \methodShort across diverse tasks, including knowledge-intensive QA and agentic tool use, proving that knowledge manipulation and execution skills can be effectively decoupled and transferred to enable robust adaptation in data-scarce target domains.
\end{itemize}

%% file: src/related_work.tex
\section{Related Work}
\label{sec:related_work}

\subsection{Knowledge Updating}
Many works aim to inject new knowledge into pre-trained Large Language Models (LLMs) by directly updating their parameters. Some works~\citep{meng2022locating,meng2023massediting} seek to precisely locate and modify specific neurons or weight matrices responsible for storing entity relationships. Others focus on text-based adaptation, where meaningful implications or synthetic Question-Answer (QA) pairs are generated from new documents to fine-tune the models~\citep{yehudai2024achieving,lampinen2025generalization,mao2025lift}. SEAL~\citep{zweiger2025self} advanced this direction by optimizing the generation of self-editing data through meta-training. Our work generally follows the latter paradigm, while advancing previous methods with a critical insight: beyond merely improving fine-tuning data, enabling the model to effectively utilize the injected knowledge is more important.

\subsection{Reinforcement Learning for LLMs}
Reinforcement learning (RL) is a critical post-training paradigm for LLMs. Recent advances, such as RL with verifiable rewards, have demonstrated the ability to elicit reasoning behaviors: DeepSeekMath introduces GRPO and improves mathematical reasoning~\citep{shao2024deepseekmath}, and DeepSeek-R1 further demonstrates that large-scale RL can yield strong reasoning capability with only limited cold-start~\citep{guo2025deepseek}. Beyond single-turn reasoning, end-to-end RL is increasingly explored for training agentic LLMs that must plan over multi-turn interactions in external environments, including web search~\citep{wei2025webagent} and tool-use agents~\citep{qian2025toolrl}. Recent analyses also suggest that RL induces favorable update dynamics---e.g., parameter updates concentrate in relatively small subnetworks~\citep{mukherjee2025reinforcement}, generalize better than SFT under distribution shift~\citep{chu2025sft}, and exhibit reduced catastrophic forgetting~\citep{shenfeld2025rl}. These properties make RL a natural candidate for injecting reusable procedural skills; however, the need for on-policy rollouts makes RL expensive to rerun for each knowledge update. We aim to reconcile this conflict: keeping RL's benefits without sacrificing the efficiency required for continual adaptation.

\subsection{Task Vectors}
Recent studies on task arithmetic view fine-tuning updates as vectors in weight space that can be composed to transfer capabilities. Concretely, a task vector is the parameter delta between a fine-tuned model and its base model, and can be added or subtracted to steer model behavior \cite{ilharco2022editing}. Building on this idea, some works \cite{du2025knowledge,cao2025param} treat such deltas as modular ``patches'' to transplant instruction-following or other reusable skills across compatible checkpoints without full fine-tuning. The closest prior work is \emph{Reasoning Vectors}~\citep{zbeeb2025reasoning}, which also extracts a residual between SFT and RL checkpoints to transfer reasoning. PaST differs from it in two essential aspects.
\textbf{(1) Theoretical assumption.} \citet{zbeeb2025reasoning} assume that reasoning capability is captured by the parameter difference between \emph{parallel} RL and SFT branches trained on the same data, but provide no formal grounding that the extracted vector is transferable or linearly composable with other parameter updates. PaST, in contrast, defines the reasoning skill as the RL-induced parameter increment on top of a knowledge-adapted SFT model, and we provide direct empirical evidence that $\Delta W_{\text{SFT}}$ and $\Delta W_{\text{RL}}$ are nearly orthogonal (Sec.~\ref{subsec:orthogonality}).
\textbf{(2) Methodology.} The above assumption leads to two concrete methodological differences. (i) \emph{Parallel vs.\ sequential extraction}: \citet{zbeeb2025reasoning} train SFT and RL independently from the same base $\theta_{\text{base}}$ and take $v = \theta_{\text{RL}} - \theta_{\text{SFT}}$, while we train sequentially within a source domain. (ii) \emph{Iterative refinement}: we further apply a multi-round refinement strategy (Sec.~\ref{subsec:iterative}) that accumulates RL-induced residuals across shifting source contexts, driving $v_{\text{skill}}$ toward content-invariant procedural logic rather than data-specific artifacts.
As \citet{cheng2023adapting} observe, training directly on raw domain corpora effectively injects factual knowledge but often impairs the model's prompting ability for question answering. Our work shows that composing RL-derived skill vectors with test-time SFT updates strengthens the model's ability to use newly incorporated knowledge for question answering.

%% file: src/motivation.tex
\section{Motivation}
\label{sec:motivation}

\subsection{The Disconnect Between Knowledge and Reasoning}
\label{subsec:disconnect}

Current adaptation methods~\citep{yehudai2024achieving,mao2025lift} predominantly rely on SFT to introduce new domain data. While SFT effectively lowers perplexity on domain documents, we hypothesize that it often fails to instill the execution logic required to manipulate that knowledge. As a result, models may “know” the facts (e.g., document content) without being able to dynamically utilize them, especially in complex settings.

To visualize this, we compare a standard Target SFT model against a Skill-Injected model (adapted using our proposed \methodShort framework, detailed in Section~\ref{sec:method}). We present a case study (Figure~\ref{fig:hero} (left), full execution trajectory along with additional case studies is provided in Appendix~\ref{app:case_study}) on a Closed-Book Tool Use task~\citep{schick2023toolformer, li2023api}, where the model must rely entirely on its parametric memory to recall API usage given only the API names.
In this instance, the user requests to download an Instagram post, but the target account is private, triggering an API error.
The SFT model correctly recalls the API name, but its reasoning collapses upon encountering the error, leading to the hallucination of non-existent tools.
In contrast, the Skill-Injected model demonstrates robust execution logic despite sharing the same knowledge base.
This observation highlights that knowledge storage (via SFT) and knowledge manipulation (via RL) are distinct capabilities. SFT alone anchors the model in the domain semantics but leaves it functionally fragile. This necessitates a method to explicitly inject robust manipulation patterns—precisely the role of our Skill Vector.

\subsection{Orthogonality of Parameter Updates}
\label{subsec:orthogonality}
While the behavioral analysis in Section~\ref{subsec:disconnect} highlights the functional difference between SFT and RL models, a fundamental question remains regarding their internal mechanics: Do knowledge acquisition and skill learning interfere with each other in the parameter space?

\begin{figure}[t]
  \includegraphics[width=\columnwidth]{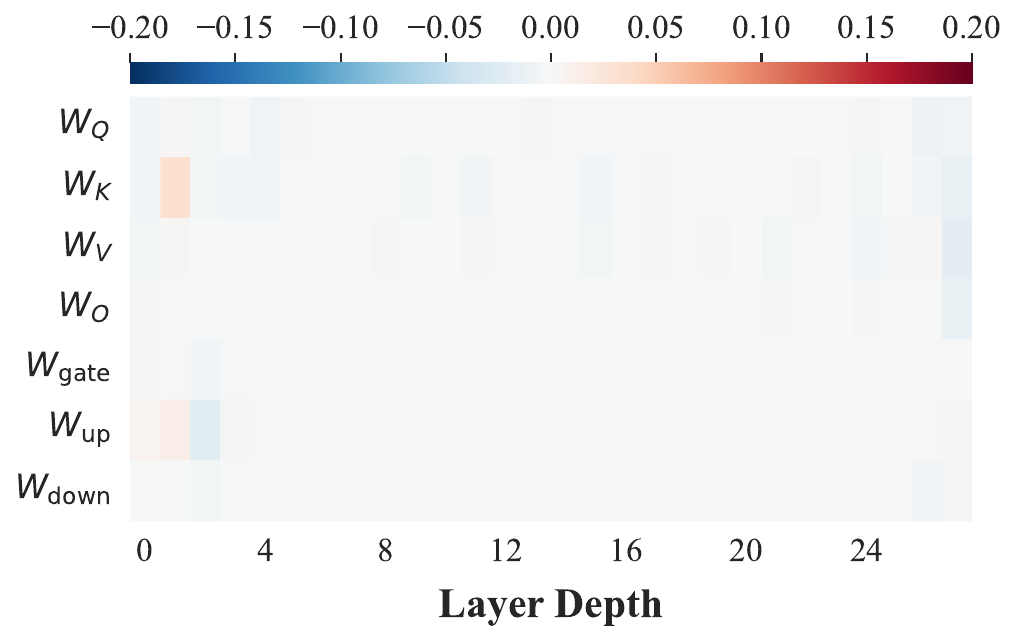}
  \caption{We visualize the layer-wise cosine similarity between the weight changes induced by SFT ($\Delta W_{\text{SFT}}$) and RL ($\Delta W_{\text{RL}}$) on the LooGLE task. The dominant near-zero values indicate that knowledge acquisition and skill learning modify the model parameters along nearly orthogonal subspaces.}
  \label{fig:orthogonality}
\end{figure}

To answer this, we analyze the weight updates. We utilize 5 documents from the LooGLE dataset to train a model sequentially via SFT and GRPO (refer to Section~\ref{subsec:loogle} for settings), and then compute the layer-wise cosine similarity between the parameter update matrices induced by each stage.
Figure~\ref{fig:orthogonality} visualizes the cosine similarity across all layers and modules. We observe a consistent trend: the correlation between
$\Delta W_{\text{SFT}}$ and $\Delta W_{\text{RL}}$ is remarkably close to zero across almost all depths and components (in contrast, two independent SFT updates remain markedly correlated; see Figure~\ref{fig:sft_sft_control} in Appendix~\ref{app:additional_vis}).
This orthogonality provides strong evidence that knowledge and manipulation skills correspond to disentangled subspaces within the high-dimensional parameter landscape.
Recent theoretical work corroborates this picture: \citet{zhu2025path} show that SFT predominantly modifies the \emph{principal} weight directions used for knowledge injection, while RLVR updates are steered by the implicit KL leash and the model geometry into the complementary \emph{off-principal} subspace. Our near-zero cosine similarity is precisely the macroscopic signature of this disjoint subspace optimization.

To understand why this parameter-level property ensures the coexistence of skills and knowledge, we analyze the signal propagation in the activation space.
As detailed in Appendix~\ref{app:proof}, assuming the input activations $x$ follow a quasi-isotropic distribution (facilitated by \verb|LayerNorm|, \citet{ba2016layer}), the expected inner product of the signals generated by these updates approximates the inner product of the weight matrices $\langle \Delta W_{\text{SFT}}, \Delta W_{\text{RL}} \rangle_F$.
High-dimensional concentration of measure~\citep{vershynin2018high} further ensures that this overlap remains minimal for individual inputs.
Given our empirical finding of near-orthogonal updates, knowledge and skill signal remain functionally disentangled, preventing destructive interference during inference and allowing downstream components to route these streams distinctively~\citep{elhage2022toy}.

This finding implies that the ``manipulation skill’’ acquired during RL is separable from the domain-specific knowledge learned via SFT, existing as an independent and extractable parameter vector $\Delta W_{\text{RL}}$. This separability motivates our approach: the skill vector can be extracted from a source domain and transferred to a target domain, enabling efficient adaptation without target-side RL.

%% file: src/method.tex
\section{\method}
\label{sec:method}

Leveraging the theoretical insight of \textit{parameter orthogonality} established in Section~\ref{subsec:orthogonality}, we propose \textbf{\methodFull}, a framework that explicitly disentangles and recombines knowledge and skills. As visualized in the right panel of Figure~\ref{fig:hero}, our approach treats the skill component as a portable vector extracted from a source domain and linearly injected into a target knowledge base.

\subsection{Problem Formulation}
\label{subsec:problem}

We consider a knowledge updating scenario involving a Source Domain $\mathcal{D}_S = \{\mathcal{C}_S, \mathcal{T}_S\}$ and a Target Domain $\mathcal{D}_T = \{\mathcal{C}_T\}$.
Here, $\mathcal{C}$ denotes unstructured knowledge documents and $\mathcal{T} = \{(x, y)\}$ represents a set of successful interaction pairs that exemplify the desired task-solving behaviors. 
While $\mathcal{D}_S$ is enriched with both knowledge and behavioral demonstrations, $\mathcal{D}_T$ contains only raw documents without task-specific labels. Our goal is to derive a target policy $\pi_{\text{target}}$ that maximizes performance on $\mathcal{D}_T$ by leveraging the skills from $\mathcal{T}_S$ and the knowledge in $\mathcal{C}_T$, eliminating the need for expensive on-policy exploration in the target domain.

\subsection{Methodology: Decoupled Skill Transfer}
\label{subsec:method}

\paragraph{Stage I: Source Skill Distillation.}
We first anchor the base model $\theta_{\text{base}}$ to the source knowledge by fine-tuning on a corpus $\mathcal{C}_S$, yielding $\theta_{S}^{\text{sft}}$. Subsequently, we apply Reinforcement Learning on trajectories $\mathcal{T}_S$ to internalize reasoning policies, resulting in $\theta_{S}^{\text{rl}}$. Leveraging our finding that RL updates occupy a subspace orthogonal to the knowledge manifold (Sec.~\ref{subsec:orthogonality}), we isolate the procedural expertise by extracting the \textbf{Skill Vector}: $\mathbf{v}_{\text{skill}} = \theta_{S}^{\text{rl}} - \theta_{S}^{\text{sft}}$. This subtraction neutralizes domain-specific declarative patterns while retaining the sparse parameter residuals responsible for internal knowledge manipulation capabilities.

\paragraph{Stage II: Target Adaptation via Vector Composition.}
To adapt to the target domain without expensive on-policy RL, we adopt a ``Compose-and-Go'' strategy. We first perform lightweight SFT on target-specific documents $\mathcal{C}_T$ to obtain $\theta_{T}^{\text{sft}}$. While this model captures target facts, it lacks the necessary reasoning logic. We then inject the source-distilled skills directly into the target parameters as $\theta_{\text{final}} = \theta_{T}^{\text{sft}} + \lambda \cdot \mathbf{v}_{\text{skill}}$, where $\lambda$ is a scaling coefficient (set to 1 in all experiments for simplicity). This linear composition grafts the source-learned reasoning geometry onto the target knowledge manifold, enabling zero-shot execution of complex tasks in the new domains.

\subsection{Iterative Skill Refinement}
\label{subsec:iterative}

A potential risk in single-round extraction is that the skill vector might overfit to the specific content distribution of the sampled source data, rather than capturing purely domain-agnostic reasoning patterns. To mitigate this, we propose an Iterative Bootstrapping Strategy.
We partition $\mathcal{D}_S$ into $K$ disjoint subsets $\{\mathcal{D}_S^{(k)}\}_{k=1}^K$ and refine the vector iteratively. 
For each round $k$, we first obtain $\theta_{S, k}^{\text{sft}}$ via SFT on the current subset, then initialize the model for reinforcement learning as $\theta_{\text{init}, k} = \theta_{S, k}^{\text{sft}} + \mathbf{v}_{k-1}$ (where $\mathbf{v}_{k-1}$ is the skill vector from the last round, and $\mathbf{v}_0 = \mathbf{0}$). The purpose is to inject the previously extracted skills as a warm-start for the RL in this round. Subsequent RL training on $\mathcal{D}_S^{(k)}$ yields $\theta_{S, k}^{\text{rl}}$, from which we extract the updated skill vector as the residual $\mathbf{v}_k = \theta_{S, k}^{\text{rl}} - \theta_{S, k}^{\text{sft}}$.
This iterative process ensures that the skill vector is progressively refined across diverse knowledge contexts, forcing the optimization to converge towards content-invariant. We empirically validate the benefit of this strategy in Section~\ref{subsubsec:ablation_iterative}.

%% file: src/experiments.tex
\section{Experiments}
\label{sec:experiments}

We empirically evaluate \methodShort across two distinct tasks: \textbf{Knowledge-based QA}~\citep{roberts2020much} and \textbf{Agentic Tool Use}~\citep{li2023api}. 
Our experiments are designed to verify: (1) \textbf{Effectiveness} on standard benchmarks (SQuAD, \citet{rajpurkar2016squad}) against strong baselines like SEAL~\citep{zweiger2025self}; (2) \textbf{Scalability} to complex, long-context scenarios (LooGLE, \citet{li2024loogle}); and (3) \textbf{Generalization} to RL-unseen tool categories (ToolBench, \citet{qin2023toolllm, guo2024stabletoolbench}).
Section~\ref{subsec:qa} details the QA results, while Section~\ref{subsec:toolbench} presents the cross-domain tool-use evaluation.
Finally, in Section~\ref{subsec:ablation}, we conduct ablation studies to analyze the impact of our \textbf{iterative skill refinement strategy} and validate the architectural necessity of our post-hoc injection method by comparing it against alternative transfer paradigms.

\subsection{Knowledge-based Question Answering}
\label{subsec:qa}

\subsubsection{Knowledge Incorporation on SQuAD}

To investigate the effectiveness of our proposed framework, we benchmark \methodShort on the task of Knowledge Incorporation using the SQuAD dataset. Unlike the original SQuAD evaluation where the passage is provided alongside the question, we adopt the closed-book setting from SEAL~\citep{zweiger2025self}. This task requires the model to first memorize the specific passage through test-time weight updates and then answer downstream questions by retrieving facts directly from its new parameter state, rather than reading them from the context window.

We utilize \verb|Qwen2.5-7B|~\citep{qwen2025qwen25technicalreport} as our base model and compare against a comprehensive set of baselines shown originally in SEAL: (1) Base Model (zero-shot); (2) Passage-Only SFT (standard fine-tuning); (3) SFT with Synthetic Data (augmenting passages with model-generated implications); (4) SFT with GPT-4.1 Data; and (5) SEAL, the current state-of-the-art method that integrates knowledge via \textit{self-edits} generated by meta-trained models. We evaluate performance under three different regimes: Single Passage updating, Continued Pretraining (CPT) on $n=200$ documents, and large-scale CPT on the full validation set ($n=2067$). 

Following the Iterative Skill Refinement strategy in Section~\ref{subsec:iterative}, we conduct two rounds of training (using the same training data as SEAL) and extract the skill vector via the parameter residual between the final RL-tuned model and its SFT counterpart. We use the same SFT data generation paradigm as the "Train on Passage + Synthetic" baseline for both Source Skill Distillation and Target Adaptation; it serves as our direct baseline to test the skill vector's contribution. The RL phase utilizes GRPO~\citep{shao2024deepseekmath} with GPT-4.1 as the reward evaluator. Detailed training configurations are provided in Appendix~\ref{app:squad}.

As shown in Table~\ref{tab:squad_results}, standard SFT on passages (33.5\%) confirms that raw text training is insufficient for knowledge retention.
By injecting our skill vector onto the "Train on Passage + Synthetic" baseline (39.7\%), \textbf{performance surges to 56.9\%}. This \textbf{+17.2\%} absolute improvement demonstrates the contribution of the skill vector, proving that while synthetic data provides the factual content, the skill vector provides the essential inference logic for answering questions accurately.
Notably, \textbf{\methodShort significantly outperforms both SEAL (47.0\%) and GPT-4.1 (46.3\%)}. While SEAL focuses on synthesizing higher-quality training data (e.g., self-edits or implications) through costly meta-training, our method achieves superior results by directly transferring intrinsic procedural skills. This suggests that the bottleneck in knowledge incorporation may not be the quality of the SFT data itself, but rather the model's underlying ability to utilize its incorporated knowledge. This trend holds in Continued Pretraining settings ($n=200/2067$), where \methodShort consistently achieves superior performance over SEAL. These results demonstrate that our skill-centric transfer paradigm is robust and scalable; it does not degrade when the underlying knowledge base expands from a single document to hundreds or thousands.

\input{tables/squad_results}

\subsubsection{Scalability to Long-Context Reasoning}
\label{subsec:loogle}

While SQuAD validates the efficacy of our method on standard-length paragraphs, real-world adaptation often requires processing extensive documentation where reasoning is complicated by the sheer volume of information. To evaluate the scalability of our framework, we conduct experiments on LooGLE, a benchmark designed for long-context understanding, consisting of realistic documents with an average length exceeding 21k tokens. 

Using \verb|Qwen2.5-7B-Instruct| as the base model, we construct a source set using the last 10 documents from the LooGLE Short Dependency QA dataset, while reserving the first 50 documents exclusively for evaluation. 
Using the source set, we perform 2 rounds of iterative skill acquisition.
In each round, we sample a batch of 5 documents and apply a specialized two-stage SFT curriculum to enforce deep knowledge encoding: (1) context memorization via multi-task training (text modeling, expansion and compression) and (2) synthetic QA training.
This is followed by GRPO to distill the retrieval logic. Full details are in Appendix~\ref{app:loogle}.

As shown in Table~\ref{tab:loogle_results}, applying the skill vector yields a significant improvement over standard Target SFT baseline which is trained using the same two-stage curriculum.
Injecting the skill vector extracted from just 5 source documents (Round 1) immediately boosts accuracy to \textbf{35.0\% (+4.9\%)}, while \textbf{the Round 2 elevates performance to 38.1\%, resulting in a cumulative gain of +8.0\%}.
These results confirm that our "how-to-recall" skill is highly transferable and successfully mitigates hallucination by transforming the model from a passive container of facts into a focused expert capable of precise parametric retrieval.

\input{tables/loogle_results}

\subsection{Cross-Domain Generalization in Tool Use}
\label{subsec:toolbench}

\begin{figure*}[ht]
    \centering
    \vspace{-2mm}
    \includegraphics[width=\textwidth]{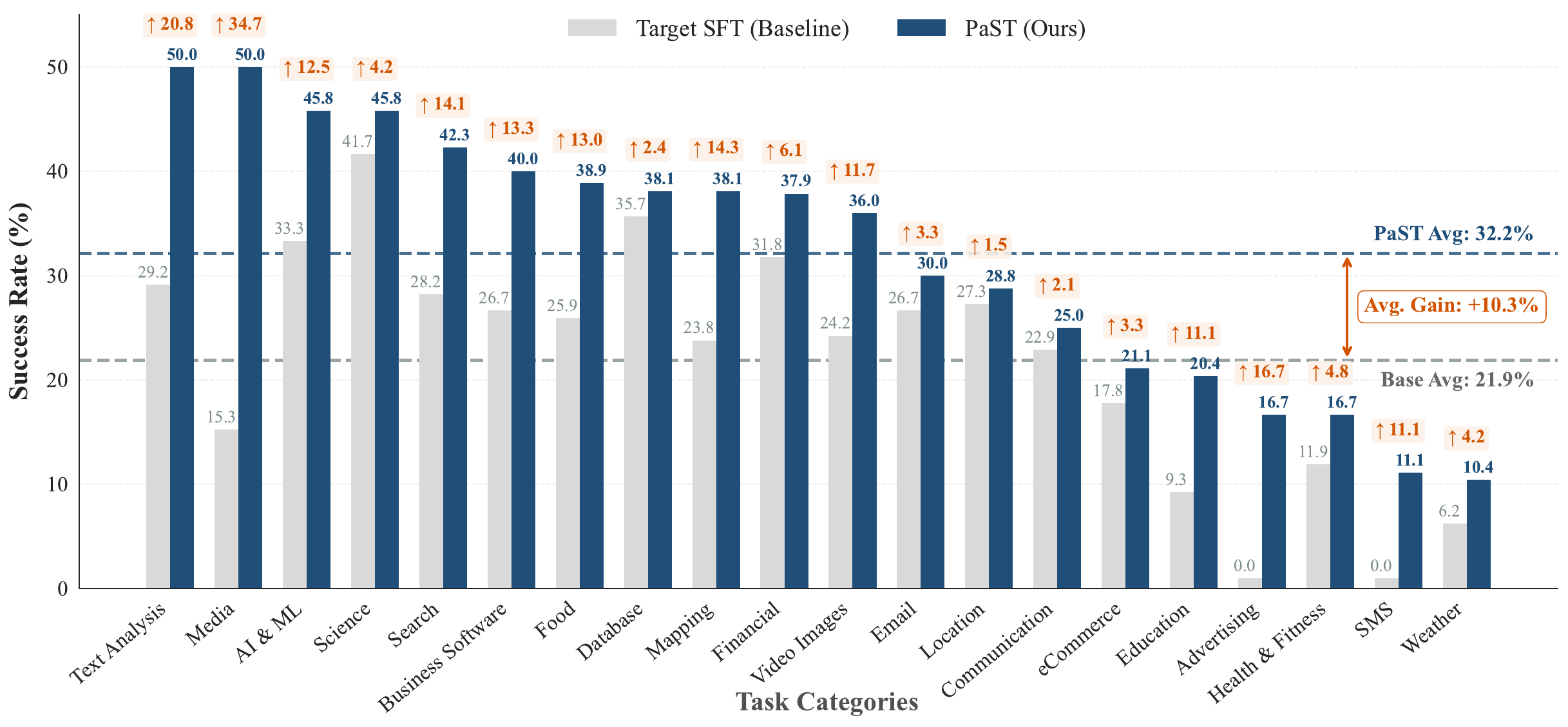}
    \vspace{-7mm}
    \caption{\textbf{Zero-shot cross-domain generalization on StableToolBench.}
    Success Rate across 20 RL-unseen target categories using a skill vector trained solely on \textit{Movies}.
    \methodShort (dark blue) raises the average success rate by \textbf{+10.3\%} over the Target SFT baseline (grey). All results are averaged over three independent runs.}
    \label{fig:toolbench_results}
    \vspace{-5mm}
\end{figure*}

We extend our evaluation to the domain of Agentic Tool Use, a task requiring the model to precisely index and utilize internalized API schemas. Unlike QA, this necessitates a mechanism of accurate parametric retrieval and multi-round execution, which we hypothesize is domain-agnostic and transferable to other tool categories.

\paragraph{Task Definition: Closed-Book Execution.}
Standard tool-use evaluations often provide full API documentation within the context window~\citep{li2023api}. However, for massive libraries with thousands of APIs, retrieving and injecting complete schemas into the context is computationally prohibitive and introduces high inference latency and token costs. To simulate this realistic constraint, we adopt a \textbf{Closed-Book Execution} setting:  the model is provided only with the names of the APIs, devoid of their detailed parameter definitions or descriptions.

\paragraph{Dataset Construction and Split.}
We utilize ToolBench, featuring 3,451 tools and 16,000+ APIs spanning 50 distinct categories.
We designate a single representative category, \textit{Movies}, as the Source Domain for skill acquisition due to its representative complexity and data richness. For the Target Domain, we identify 20 distinct categories entirely unseen during RL, filtered for data sufficiency and API count to ensure a balanced evaluation (see Appendix~\ref{app:toolbench} for details). For testing, we utilize the curated solvable queries from StableToolBench, covering both single-tool (G1) and intra-category multi-tool (G2) scenarios.

\paragraph{Training Setup.}
Using \verb|Qwen2.5-7B-Instruct|, we first establish a robust mapping between API names and functionalities on source domain via SFT on a composite dataset: raw API schemas, natural language transcriptions, and bidirectional QA pairs (training the model to predict usage from API names and vice versa). 
Additionally, we perform a format-alignment SFT using the initial steps of the trajectories in ToolBench data to instill the ReAct convention.
To refine the execution policy, we employ PPO \citep{schulman2017proximal} implemented by adapting the Search-R1 \citep{jin2025search} framework.
During training, we employ GPT-4o-mini as an environment simulator to generate realistic API return values.
The reward signal is a composite of format rewards (JSON syntax and ReAct format), execution rewards (successful API returns), and solution rewards (judged by GPT-4.1 for intent resolution).
Details are provided in Appendix~\ref{app:toolbench}.

\paragraph{Target Adaptation and Results.}
Following the two-stage adaptation (SFT as described above, followed by skill vector injection), \methodShort significantly outperforms the Target SFT baseline.
As visualized in Figure~\ref{fig:toolbench_results}, \textbf{our method increases the average success rate from 21.9\% to 32.2\%.} Notably, \methodShort achieves zero-shot activation in domains where the baseline fails completely (\textit{Advertising} 0\% $\to$ 16.7\% and \textit{SMS} 0\% $\to$ 11.1\%).
Remarkably, \methodShort outperforms the baseline in all 20 evaluated categories, demonstrating robust positive transfer across diverse domains.

\subsection{Ablation Studies}
\label{subsec:ablation}

\subsubsection{Impact of Iterative Skill Refinement}
\label{subsubsec:ablation_iterative}

We first validate the effectiveness of the \textbf{Iterative Skill Refinement} by comparing it against Single-Round baselines on SQuAD, maintaining equal total optimization steps and data volume. As shown in Table~\ref{tab:ablation_iterative}, simply doubling the source data in a single round ($N=100$) often yields marginal gains or even performance degradation, suggesting that reasoning logic becomes overfitted to specific source content. In contrast, our Iterative strategy ($K=2$, $N=50$ per round) consistently achieves the highest accuracy. This confirms that iterative refinement forces the skill vector to capture content-invariant execution logic, preventing the reasoning policy from being inextricably bound to a specific set of source facts.

\input{tables/ablation_iterative}

\paragraph{$K \times M$ trade-off.}
To further understand how the sharding granularity affects the extracted skill, we conduct a finer-grained sweep on LooGLE that fixes the total source documents ($K \cdot M = 10$) and the total RL optimization steps, while varying the number of rounds $K$ and documents per round $M$ (Table~\ref{tab:ablation_kxm}). Increasing $K$ from $1$ to $5$ steadily improves accuracy ($42.9 \!\to\! 45.3$) by exposing the RL phase to shifting knowledge backgrounds, which acts as regularization that filters out content-specific artifacts. Pushing too far ($10 \times 1$) hurts: a single-document shard yields a narrow reward distribution and high gradient variance, preventing stable convergence within the limited per-round step budget. In practice, $K$ should be increased to promote decoupling, while keeping each shard sufficiently representative for a stable RL signal.

\input{tables/ablation_kxm}

\subsubsection{Impact of Transfer Strategy}
\label{subsubsec:ablation_strategy}

Finally, we analyze the optimal stage for skill injection by comparing our Post-hoc Composition against two alternative paradigms on the LooGLE benchmark: (1) \textbf{Sequential Fine-Tuning}, where the source RL model $\theta_{S}^{\text{rl}}$ is directly fine-tuned on target documents; and (2) \textbf{Pre-Injection}, where $\mathbf{v}_{\text{skill}}$ is added to $\theta_{\text{base}}$ before target SFT. 
Following the setup in Section~\ref{subsec:loogle}, we report the results on the first 10 documents of the test set in Table~\ref{tab:ablation_strategy}.
Interestingly, Sequential Fine-Tuning (30.3) performs slightly worse than the standard Target SFT baseline (32.9). This suggests that directly optimizing for new knowledge on top of RL parameters may induce optimization conflicts that potentially disrupt the delicate reasoning circuitry learned during RL.
Pre-Injection achieves moderate performance (36.5) but lags behind our method, likely because subsequent SFT shifts the weight manifold and misaligns the pre-injected skills.
Post-hoc Composition (Ours) yields the highest accuracy (44.6) by anchoring the declarative knowledge first, ensuring execution logic is grafted onto a stable knowledge representation without being distorted by the SFT optimization trajectory.

\input{tables/ablation_strategy}

\subsubsection{Comparison with Target-Domain RL}
\label{subsubsec:target_rl_compare}

A natural concern is whether \methodShort sacrifices substantial accuracy by skipping target-domain RL altogether. To quantify this gap, we directly compare \methodShort against running GRPO on the target domain at varying step budgets, on the first 10 LooGLE documents (the same setting as Sec.~\ref{subsubsec:ablation_strategy}). All target RL runs use the same hyperparameters as in Sec.~\ref{subsec:loogle} and are timed on 8$\times$A100 GPUs.
As shown in Table~\ref{tab:target_rl_compare}, \methodShort matches the accuracy of target RL trained for $\sim$75 steps and trails 100-step target RL by only $1.6$ points, while incurring \textbf{zero} additional training time on the target domain. Even modest target RL (75 steps) costs $72$ minutes \emph{per document}, which scales prohibitively under continual adaptation. This confirms that the skill vector recovers most of the benefit of target RL at a tiny fraction of its cost.

\input{tables/target_rl_compare}

%% file: tables/squad_results.tex
\begin{table*}
    \centering
    \caption{Mean accuracy on SQuAD (no-context) across different adaptation regimes. Values for baselines are taken from SEAL~\citep{zweiger2025self}. The values in parentheses denote the absolute improvement over the ``Train on Passage + Synthetic'' baseline.}
    \begin{tabular}{lccc}
    \toprule
    \multirow{2}{*}{\textbf{Method}} & \textbf{Single Passage} & \textbf{CPT} & \textbf{CPT} \\
    & (n = 1; LoRA) & (n = 200; full-FT) & (n = 2067; full-FT) \\
    \midrule
    Base Model & 32.7 & 32.7 & 29.0 \\
    Train on Passage & 33.5 & 36.0 & 31.2 \\
    Train on Passage + Synthetic & 39.7 & 50.6 & 43.4 \\
    Train on Passage + GPT-4.1 Synthetic & 46.3 & \textbf{59.4} & \textbf{49.2} \\
    SEAL & 47.0 & 58.2 & 46.4 \\
    \midrule
    \rowcolor{gray!10} \textbf{\methodShort 50 (Ours)} & \underline{50.8}\gain{11.1} & \underline{58.9}\gain{8.3} & \underline{47.4}\gain{4.0} \\
    \rowcolor{gray!10} \textbf{\methodShort 50 $\times$ 2 (Ours)} & \textbf{56.9}\gain{17.2} & 58.7\gain{8.1} & \textbf{49.2}\gain{5.8} \\
    \bottomrule
    \end{tabular}
    \label{tab:squad_results}
\end{table*}

%% file: tables/loogle_results.tex
\begin{table}[ht]
\centering
\caption{\textbf{Long-Context QA Performance on LooGLE (Short Dependence QA).} Comparing standard adaptation methods against \methodShort, the Skill Vector significantly enhances the model's ability to retrieve and reason over extremely massive information. All results are averaged over three independent runs.}
\begin{tabular}{lc}
\toprule
\textbf{Method} & \textbf{Accuracy} \\
\midrule
SFT & 30.1 \\
\textbf{\methodShort 5} & \underline{35.0}\gain{4.9} \\
\textbf{\methodShort 5 $\times$ 2} & \textbf{38.1}\gain{8.0} \\
\bottomrule
\end{tabular}
\label{tab:loogle_results}
\end{table}

%% file: tables/ablation_iterative.tex
\begin{table}[ht]
    \centering
    \small
    \caption{\textbf{Ablation on Iterative Refinement (SQuAD).} We compare Single-Round training (with half and full data) against our Iterative strategy on SQuAD. $N$ denotes the number of \textit{source training documents} in each round.}
    \label{tab:ablation_iterative}
    \resizebox{\columnwidth}{!}{
    \begin{tabular}{lccc}
        \toprule
        \multirow{2}{1.5cm}[-2.5pt]{\centering \textbf{Training Rounds}} & \multicolumn{3}{c}{\textbf{SQuAD}} \\
        \cmidrule(lr){2-4}
         & Single & CPT (200) & CPT (2K) \\
        \midrule
        1 ($N=50$) & 50.8 & \textbf{58.9} & 47.4 \\
        1 ($N=100$) & 49.9 & 58.3 & 47.1 \\
        \textbf{2 ($N=50$)} & \textbf{56.9} & 58.7 & \textbf{49.2} \\
        \bottomrule
    \end{tabular}
    }
\end{table}

%% file: tables/ablation_kxm.tex
\begin{table}[ht]
    \centering
    \small
    \caption{\textbf{$K \times M$ Trade-off on LooGLE.} Fixing the total source documents ($K \cdot M = 10$) and total RL optimization steps, we sweep the number of rounds $K$ versus documents per round $M$.}
    \label{tab:ablation_kxm}
    \begin{tabular}{lc}
        \toprule
        \textbf{Rounds $K$ $\times$ Docs/Round $M$} & \textbf{LooGLE Acc} \\
        \midrule
        $1 \times 10$ (Single-round) & 42.9 \\
        $2 \times 5$                 & 44.6 \\
        $\mathbf{5 \times 2}$        & \textbf{45.3} \\
        $10 \times 1$                & 42.1 \\
        \bottomrule
    \end{tabular}
\end{table}

%% file: tables/ablation_strategy.tex
\begin{table}[ht]
    \centering
    \small
    \caption{\textbf{Ablation on Transfer Strategy.} We evaluate different methods of combining source skills with target knowledge on LooGLE. ``Post-hoc Composition'' (Ours) significantly outperforms sequential training or pre-injection methods.}
    \label{tab:ablation_strategy}
    \begin{tabular}{lc}
        \toprule
        \textbf{Transfer Strategy} & \textbf{LooGLE Accuracy} \\
        \midrule
        Target SFT & 32.9 \\
        Sequential FT & 30.3 \\
        Pre-Injection & 36.5 \\
        \textbf{Post-hoc Composition} & \textbf{44.6} \\
        \bottomrule
    \end{tabular}
\end{table}

%% file: tables/target_rl_compare.tex
\begin{table}[ht]
    \centering
    \small
    \caption{\textbf{Comparison with Target-Domain RL on LooGLE.} We compare \methodShort against direct GRPO trained on the target domain at varying step budgets, on the first 10 LooGLE documents. Runtime measures the per-document target-side RL training time on 8$\times$A100 GPUs.}
    \label{tab:target_rl_compare}
    \begin{tabular}{lcc}
        \toprule
        \textbf{Method} & \textbf{Acc} & \textbf{Test-time RL Time} \\
        \midrule
        Target SFT             & 32.9          & 0 min \\
        Target RL (25 steps)   & 38.2          & 33 min \\
        Target RL (50 steps)   & 41.2          & 52 min \\
        Target RL (75 steps)   & 44.0          & 72 min \\
        Target RL (100 steps)  & \textbf{46.2} & 91 min \\
        \textbf{\methodShort (Ours)} & \textbf{44.6} & \textbf{0 min} \\
        \bottomrule
    \end{tabular}
\end{table}

%% file: src/conclusion.tex
\section{Conclusion}
\label{sec:conclusion}

In this paper, we introduced \methodFull to bridge the functional disconnect between knowledge acquisition and reasoning skills in LLMs. By identifying that SFT and RL parameter updates are nearly orthogonal, we developed a modular framework to extract a domain-agnostic Skill Vector ($v_{skill}$) from source tasks and linearly inject it into models adapted to new data. Our evaluations on SQUAD, LooGLE, and ToolBench demonstrate that PaST significantly enhances a model’s ability to manipulate newly internalized knowledge. Ultimately, \methodShort offers a computationally efficient and scalable alternative to on-policy RL, enabling effective knowledge adaptation.

%% file: src/limitation.tex
\section*{Limitations}
\label{sec:limitation}

Despite the effectiveness of \methodShort, several limitations remain to be addressed in future work:
\begin{itemize}
    \item Breadth of Experimental Domains: While we evaluated our framework on standard QA and agentic tool-use benchmarks , the diversity of "source-to-target" transfer scenarios could be further expanded. 
    \item Static Scaling Coefficient: For simplicity, the scaling coefficient  in our injection formula ($\theta_{final} = \theta_{T}^{sft} + \lambda \cdot v_{skill}$) was consistently set to 1 across all experiments. However, we hypothesize that the optimal $\lambda$ might vary depending on the gap between the source and target knowledge manifolds or the specific model architecture.
    \item Model Architecture Generalization: Our empirical observations and experiments were primarily conducted using the \verb|Qwen2.5-7B| and \verb|Qwen2.5-7B-Instruct|. While our theoretical proof regarding orthogonality is grounded in general properties of high-dimensional parameter spaces , additional studies are needed to confirm if these update dynamics hold consistently across a broader range of model scales and architectures.
\end{itemize}

%% file: src/acknowledgement.tex
\section*{Acknowledgments}
This work is supported by National Natural Science Foundation of China (62276003, 62561160152) and Kunpeng\&Ascend Center of Excellence, Peking University.

%% file: src/appendix/main.tex
\appendix
\clearpage

\input{src/appendix/case_study}
\input{src/appendix/sft_cosine}
\input{src/appendix/orthogonality_extended}
\input{src/appendix/proof}
\input{src/appendix/squad_settings}
\input{src/appendix/loogle_settings}
\input{src/appendix/toolbench_settings}
\input{src/appendix/ai_usage}

%% file: src/appendix/case_study.tex
\section{Detailed Case Studies}
\label{app:case_study}

\subsection{Full Execution Trajectory: Instagram Post Task}

In this section, we provide the complete interaction trace for the example discussed in Section~\ref{subsec:toolbench}. This comparison highlights the difference in reasoning logic when facing a ``Private Account'' error. The full interaction trajectories on both models are shown in Table~\ref{tab:toolbench_case1} and \ref{tab:toolbench_case2}.

\input{tables/toolbench_case}

\subsection{Parametric Knowledge Retrieval: SQuAD Case Study}

We present a comparison on the SQuAD dataset in Table~\ref{tab:squad_case}, involving a specific legal context regarding EU Directives. 
The SFT baseline suffers from knowledge fallback, ignoring the internalized document and providing a generic answer about legal liability based on its pre-training priors. 
In contrast, our model demonstrates precise parametric retrieval, successfully locating the specific term ``Directives'' within its weights and accurately synthesizing the supporting explanation (e.g., lack of ``horizontal direct effect'') as presented in the hidden context.

\input{tables/squad_case}

%% file: tables/toolbench_case.tex
\begin{table*}[t!] 
\centering
\begin{mybox}[User Query]{gray}
    \small I'm trying to download a post and a reel from Instagram. Can you provide me with the download links for the post and reel? The post link is [post link] and the reel link is [reel link].
\end{mybox}

\vspace{-2mm}

\begin{mybox}[Baseline: SFT Model]{red!50!gray}
\begin{lstlisting}[style=trajectoryStyle]
Thought: I should call the 'posts_for_instagram_reels_and_post_downloader' function with the argument 'link' set to [post link]. This will allow me to download the Instagram post URL [post link] and retrieve the download links for the reels and posts from the specified Instagram media URL.
Action: posts_for_instagram_reels_and_post_downloader
Action Input: {"link": [post link]}

Observation: {"error": "", "response": "{ Sorry but this instagram account is private }"}

Thought: The API call to download the Instagram post failed because the account is private. To proceed, I should try another approach. 
Action: newapi
Action Input: {"url": [post link]}

Observation: {"error": "Invalid API name: 'newapi'. Please check the API name.", "response": ""}

Thought: The API call failed because the API name is invalid. This indicates that the 'newapi' function is not available. To proceed, I can try using the 'tiktok_download_video' function instead.
Action: get_tiktok_video_info_for_tiktok_download_video
Action Input: {"hd": "False", "url": [post link]}

Observation: {"error": "Invalid API name: 'get_tiktok_video_info_for_tiktok_download_video'. Please check the API name.", "response": ""}

Thought: The API call failed because the API name is invalid. This suggests that the 'get_tiktok_video_info_for_tiktok_download_video' function is not available. As a last resort, I can use the 'cheap_youtube_api' to download the Instagram post.
Action: trending_music_gaming_news_movies_for_cheap_youtube_api
Action Input: {"category": "Instagram", "region": "US", "limit": 1}

Observation: {"error": "Invalid API name: 'trending_music_gaming_news_movies_for_cheap_youtube_api'. Please check the API name.", "response": ""}

Thought: Since the tool call limit has been reached, I now need to summarize my thoughts and call 'Finish' to end the task. I tried using different APIs to download the Instagram post and retrieve the download links, but none of them worked. It seems that the 'instagram_reels_and_post_downloader' function is not functioning properly. To handle this, I can provide the user with an alternative solution. I can suggest using a third-party app or website that specifically caters to downloading Instagram posts and reels. However, I should also mention that the user may need to provide the post URL and the download links separately.
Action: Finish
Action Input: {"final_answer": "The current API does not have a valid function available for downloading Instagram posts and reels. As an alternative, you can try using third-party apps or websites specifically designed for this purpose. When using these apps, make sure to provide the Instagram post URL and the download links separately. This will help you download the desired Instagram media efficiently. Sorry for the inconvenience caused by this issue."}
\end{lstlisting}
\end{mybox}
\vspace{-4mm}
\caption{Case Study: Model comparison on tool-use trajectories (Part 1). Links are masked due to privacy concerns.}
\label{tab:toolbench_case1}
\end{table*}

\begin{table*}[t!]
\centering

\begin{mybox}[Ours]{blue!50!gray}
\begin{lstlisting}[style=trajectoryStyle]
Thought: I am calling the 'posts_for_instagram_reels_and_post_downloader' function with the argument of 'link' set to [post link]. This is because the user wants to download a post from Instagram and has provided the post link. By using this function, I can retrieve the download link for the specified post.
Action: posts_for_instagram_reels_and_post_downloader
Action Input: {"link": [post link]}

Observation: {"error": "", "response": "{ Sorry but this instagram account is private }"}

Thought: The response indicates that the Instagram account is private, which is why I am unable to retrieve the download link for the post. It seems that the user needs to make the account public in order to access the download link.
Action: posts_for_instagram_reels_and_post_downloader
Action Input: {"link": [reel link]}

Observation: {"error": "", "response": "{ Sorry but this instagram account is private }"}

Thought: The response for the reel download link is the same as for the post, indicating that the Instagram account is also private. Therefore, the user needs to make the account public in order to access the download links for both the post and reel.
Action: Finish
Action Input: {"final_answer": "The Instagram account is private, so I am unable to retrieve the download links for the post and reel. The user needs to make the account public in order to access the download links."}
\end{lstlisting}
\end{mybox}
\vspace{-4mm}
\caption{Case Study: Model comparison on tool-use trajectories (Part 2). Links are masked due to privacy concerns.}
\label{tab:toolbench_case2}
\end{table*}

%% file: tables/squad_case.tex
\begin{table*}[t!]
\centering
\begin{mybox}[Context (Not shown to the model)]{gray}
    \small While the Treaties and Regulations will have direct effect (if clear, unconditional and immediate), Directives do not generally give citizens (as opposed to the member state) standing to sue other citizens. In theory, this is because TFEU article 288 says Directives are addressed to the member states and usually \"leave to the national authorities the choice of form and methods\" to implement. In part this reflects that directives often create minimum standards, leaving member states to apply higher standards. For example, the Working Time Directive requires that every worker has at least 4 weeks paid holidays each year, but most member states require more than 28 days in national law. However, on the current position adopted by the Court of Justice, citizens have standing to make claims based on national laws that implement Directives, but not from Directives themselves. Directives do not have so called \"horizontal\" direct effect (i.e. between non-state parties). This view was instantly controversial, and in the early 1990s three Advocate Generals persuasively argued that Directives should create rights and duties for all citizens. The Court of Justice refused, but there are five large exceptions.
\end{mybox}

\vspace{-2mm}

\begin{mybox}[Question]{green!50!gray}
    \small What generally does not allow citizens to sue other citizens?
\end{mybox}

\vspace{-2mm}

\begin{mybox}[Ground Truth]{green}
    \small Directives.
\end{mybox}

\begin{mybox}[Baseline: SFT Model]{red!50!gray}
    \small In general, citizens cannot directly sue other citizens in most legal systems. However, there are exceptions where private individuals can be held liable for certain types of harm, such as intentional torts or breaches of contract.
\end{mybox}

\begin{mybox}[Ours]{blue!50!gray}
    \small Directives do not create rights and duties for citizens. They are addressed to Member States and usually have indirect effect, meaning that they do not create rights and duties for citizens. Directives usually contain exemptions and allow Member States to choose the form in which they are implemented. This means that there is often no direct effect between citizens and in most cases, citizens cannot sue other citizens based on Directives. However, Directives often create higher standards than those that exist in national laws, so if a national law is less strict than a Directive, the latter can be used to argue that a law is invalid. In summary, citizens generally cannot sue other citizens based on Directives, but Directives can create higher standards than national laws, which can be used to challenge the validity of national laws.
\end{mybox}
\vspace{-4mm}
\caption{Case Study: Model comparison on SQuAD.}
\label{tab:squad_case}
\end{table*}

%% file: src/appendix/sft_cosine.tex
\section{Additional Visualization: Orthogonality Control Experiment}
\label{app:additional_vis}

In Section~\ref{sec:motivation}, we argued that the parameter updates for knowledge acquisition ($\Delta W_{\text{SFT}}$) and skill learning ($\Delta W_{\text{RL}}$) are structurally disentangled, evidenced by their near-zero cosine similarity. A potential counter-argument is that in high-dimensional parameter spaces (e.g., $d_{\text{model}} \gg 1$), random vectors naturally tend to be orthogonal.
To rule out the possibility that our observation is merely a statistical artifact of high dimensionality, we conducted a control experiment to measure the similarity between two updates of the \textit{same} modality (i.e., Knowledge vs. Knowledge).

\paragraph{Experimental Setup.}
Using the same LooGLE dataset setting as the main experiment, we performed two consecutive rounds of Supervised Fine-Tuning (SFT) on disjoint data subsets.
We then computed the layer-wise cosine similarity $\text{Sim}(\Delta W_{\text{SFT1}}, \Delta W_{\text{SFT2}})= \frac{\langle \Delta W_{\text{SFT1}}, \Delta W_{\text{SFT2}} \rangle_F}{||\Delta W_{\text{SFT1}}||_F \cdot ||\Delta W_{\text{SFT2}}||_F}$
where $\langle A, B \rangle_F = \text{Tr}(A B^\top) = \sum_{i,j} A_{ij} B_{ij}$ denotes the Frobenius inner product~\citep{golub2013matrix}.

\begin{figure}[t]
  \includegraphics[width=\columnwidth]{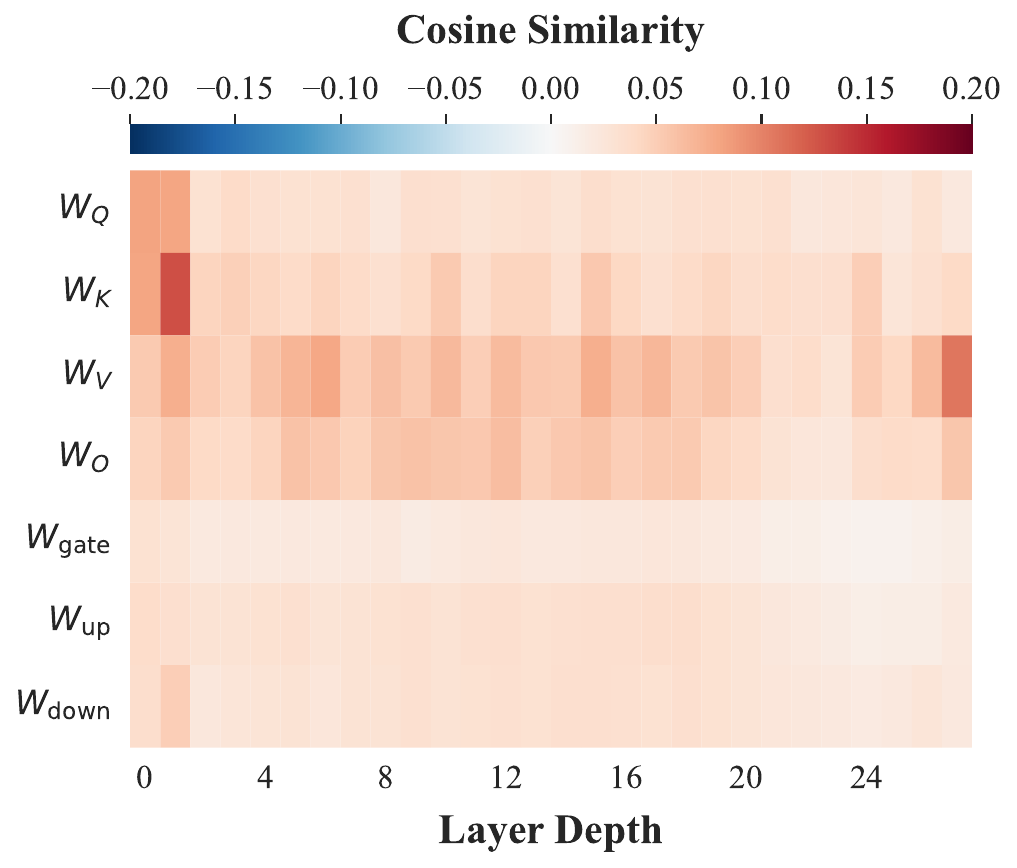}
  \caption{\textbf{Control Experiment: Similarity between two SFT updates.} We visualize the cosine similarity between parameter updates induced by two different rounds of SFT ($\Delta W_{\text{SFT21}}$ vs. $\Delta W_{\text{SFT2}}$) on LooGLE. Unlike the SFT-RL comparison, these updates show a clear positive correlation (red regions), indicating that knowledge injection tasks operate within a shared parameter subspace.}
  \label{fig:sft_sft_control}
\end{figure}

\paragraph{Result Analysis.}
Figure~\ref{fig:sft_sft_control} presents the resulting heatmap. In stark contrast to the SFT-RL comparison (Figure~\ref{fig:orthogonality} in the main text), the SFT-SFT heatmap exhibits a distinct positive correlation (indicated by the prevalent orange/red hues) across most layers.
This comparison provides two critical insights:
\begin{enumerate}
\item \textbf{Manifold Alignment of Knowledge:} Tasks of the same nature (injecting declarative facts) tend to modify the model parameters along a shared or aligned subspace, resulting in non-zero cosine similarity.
\item \textbf{Validation of Disentanglement:} The fact that $\Delta W_{\text{SFT}}$ vs. $\Delta W_{\text{SFT}}$ shows correlation while $\Delta W_{\text{RL}}$ vs. $\Delta W_{\text{SFT}}$ does not confirms that the orthogonality observed in our main result is a genuine property of the Knowledge-Skill decomposition, rather than a geometric triviality.
\end{enumerate}

%% file: src/appendix/orthogonality_extended.tex
\section{Robustness of the Orthogonality Finding}
\label{app:orthogonality_extended}

The near-zero cosine similarity between $\Delta W_{\text{SFT}}$ and $\Delta W_{\text{RL}}$ reported in Section~\ref{subsec:orthogonality} is measured in a single (LooGLE, GRPO) configuration. To support our claim that this orthogonality is a property of the optimization dynamics rather than an artifact of a specific task, RL algorithm, or reward design, we provide two extended studies below.

\subsection{Across Tasks, RL Algorithms, and Reward Designs}
\label{app:orth_cross_task}

We compute the average layer-wise cosine similarity between $\Delta W_{\text{SFT}}$ and $\Delta W_{\text{RL}}$ on three benchmarks that differ in task type, RL algorithm, and reward source. As shown in Table~\ref{tab:orth_cross_task}, the similarity remains close to zero ($\sim\!10^{-3}$) across all settings, indicating that the SFT--RL orthogonality is largely invariant to these design choices.

\begin{table}[ht]
    \centering
    \small
    \caption{\textbf{Orthogonality across tasks, RL algorithms, and rewards.} Average cosine similarity between $\Delta W_{\text{SFT}}$ and $\Delta W_{\text{RL}}$ across all model modules.}
    \label{tab:orth_cross_task}
    \resizebox{\columnwidth}{!}{
    \begin{tabular}{lccc}
        \toprule
        \textbf{Benchmark} & \textbf{RL Algo.} & \textbf{Reward} & \textbf{Avg. Cos. Sim.} \\
        \midrule
        LooGLE    & GRPO & LLM Judge      & $-0.0018$ \\
        SQuAD     & GRPO & LLM Judge      & $-0.0019$ \\
        ToolBench & PPO  & Env. Simulator & $-0.0020$ \\
        \bottomrule
    \end{tabular}
    }
\end{table}

\subsection{Cross-Domain Orthogonality}
\label{app:orth_cross_domain}

A natural follow-up question is whether the orthogonality between SFT and RL updates also holds when the two updates are obtained from \emph{different} domains (i.e., the SFT update from the source domain and the RL update from the target domain, or vice versa). Using the LooGLE source/target split from Section~\ref{subsec:loogle}, we compute the average cosine similarity for all four SFT--RL combinations, together with two same-modality controls (SFT vs.\ SFT and RL vs.\ RL across domains).

\begin{table}[ht]
    \centering
    \small
    \caption{\textbf{Cross-domain orthogonality on LooGLE.} Average cosine similarity between parameter update matrices across all model modules of \texttt{Qwen2.5-7B-Instruct}.}
    \label{tab:orth_cross_domain}
    \begin{tabular}{lc}
        \toprule
        \textbf{Comparison} & \textbf{Avg. Cos. Sim.} \\
        \midrule
        SFT (Source) vs. RL (Source)  & $-0.00180$ \\
        SFT (Source) vs. RL (Target)  & $-0.00067$ \\
        SFT (Target) vs. RL (Source)  & $-0.00109$ \\
        \midrule
        SFT (Source) vs. SFT (Target) & $\phantom{-}0.03132$ \\
        RL (Source) vs. RL (Target)   & $\phantom{-}0.02575$ \\
        \bottomrule
    \end{tabular}
\end{table}

Two observations follow from Table~\ref{tab:orth_cross_domain}. First, the SFT--RL similarity stays at the $10^{-3}$--$10^{-4}$ level even when the two updates are obtained from different domains, confirming that the orthogonality is a property of the SFT vs.\ RL optimization regimes themselves rather than of any particular dataset alignment. Second, the same-modality controls (SFT vs.\ SFT and RL vs.\ RL) are $1$--$2$ orders of magnitude larger in magnitude, ruling out the possibility that the near-zero SFT--RL value is a generic high-dimensional geometric triviality. Together with the cross-task results in Table~\ref{tab:orth_cross_task}, this strengthens the empirical basis for treating $v_{\text{skill}} = \theta^{\text{rl}} - \theta^{\text{sft}}$ as a transferable, domain-agnostic skill direction.

%% file: src/appendix/proof.tex
\section{Theoretical Proof of Functional Disentanglement}
\label{app:proof}

In Section~\ref{subsec:orthogonality}, we empirically observed that the parameter update matrices for knowledge acquisition ($\Delta W_{\text{SFT}}$) and skill learning ($\Delta W_{\text{RL}}$) are nearly orthogonal in terms of the Frobenius inner product. Here, we provide a formal derivation showing why this parameter-level orthogonality guarantees functional disentanglement in the activation space.

\subsection{Preliminaries and Assumptions}

Let $x \in \mathbb{R}^{1 \times d}$ be the input activation vector at a given layer, where $d$ is the model dimension (e.g., 4096).
Let $A = \Delta W_{\text{SFT}} \in \mathbb{R}^{d \times d}$ and $B = \Delta W_{\text{RL}} \in \mathbb{R}^{d \times d}$ be the weight update matrices.
The signals generated by these updates are $u = xA$ and $v = xB$, respectively.

We make two standard assumptions regarding the statistical properties of deep neural networks:
\begin{enumerate}
    \item \textbf{Parameter Orthogonality:} Based on our empirical observations, we assume $\langle A, B \rangle_F = \text{Tr}(AB^\top) \approx 0$.
    \item \textbf{Isotropic Inputs:} We assume the input activations $x$ are zero-centered and quasi-isotropic, with covariance proportional to the identity matrix. This is a common property in Transformers facilitated by \verb|LayerNorm| \citep{ba2016layer}:
    \begin{equation}
        \mathbb{E}[x^\top x] = \sigma^2 I
    \end{equation}
    where $\sigma^2$ is the variance of the activations.
\end{enumerate}

\subsection{Derivation of Signal Orthogonality}

We investigate the interference between the knowledge signal ($u$) and the skill signal ($v$) by examining their inner product $\langle u, v \rangle$.

\paragraph{1. Expectation of Signal Overlap.}
The expected inner product of the generated signals over the data distribution is:
\begin{align}
    \mathbb{E}[\langle u, v \rangle] = \mathbb{E}[u v^\top] &= \mathbb{E}[(xA)(xB)^\top] \nonumber \\
    &= \mathbb{E}[x A B^\top x^\top]
\end{align}
Using the property that the trace of a scalar is the scalar itself ($\text{Tr}(c) = c$) and the cyclic property of the trace ($\text{Tr}(XYZ) = \text{Tr}(YZX)$):
\begin{align}
    \mathbb{E}[x A B^\top x^\top] &= \mathbb{E}[\text{Tr}(x A B^\top x^\top)]\nonumber \\
    &= \mathbb{E}[\text{Tr}(A B^\top x^\top x)] \nonumber \\
    &= \text{Tr}(A B^\top \mathbb{E}[x^\top x])
\end{align}
Substituting the isotropic assumption $\mathbb{E}[x^\top x] = \sigma^2 I$:
\begin{equation}
    \mathbb{E}[\langle u, v \rangle] = \text{Tr}(A B^\top \cdot \sigma^2 I) = \sigma^2 \langle A, B \rangle_F
\end{equation}
\textbf{Conclusion 1:} Since $\langle A, B \rangle_F \approx 0$, the expected overlap between the knowledge and skill signals is zero ($\mathbb{E}[\langle u, v \rangle] \approx 0$).

\paragraph{2. Concentration in High Dimensions.}
While the expectation is zero, we must ensure that the variance is low enough such that the overlap is minimal for \textit{any individual input} $x$. This is guaranteed by the \textbf{Concentration of Measure} phenomenon in high-dimensional spaces \citep{vershynin2018high}.

Let $M = AB^\top$. Consider the quadratic form $Y = x M x^\top$. For a random vector $x$ with independent sub-Gaussian components (a reasonable approximation for normalized representations), the Hanson-Wright inequality bounds the deviation from the mean:
\begin{align}
    P(|\langle u&, v \rangle - \mathbb{E}[\langle u, v \rangle]| \geq t) \nonumber \\
    \leq& 2 \exp \left( -c \min \left( \frac{t^2}{\sigma^4 ||M||_F^2}, \frac{t}{\sigma^2 ||M||_2} \right) \right)
\end{align}
where $c$ is a universal constant.
In modern LLMs where $d \gg 1$, the Frobenius norm $||M||_F$ grows with $\sqrt{d}$, but the concentration probability improves exponentially. This implies that for the vast majority of input samples, the actual inner product $\langle u, v \rangle$ will be tightly concentrated around its expectation (zero).

\subsection{Functional Implication}
This derivation proves that parameter-level orthogonality ($\Delta W_{\text{SFT}} \perp \Delta W_{\text{RL}}$) translates directly to signal-level orthogonality in the activation space. Consequently, the ``knowledge signal'' and ``skill signal'' propagate through the network as functionally independent components, preventing destructive interference and enabling the subsequent layers (e.g., attention heads) to attend to them distinctively.

%% file: src/appendix/squad_settings.tex
\section{SQuAD Experiments}
\label{app:squad}

\paragraph{Task definition.}
We follow the closed-book SQuAD knowledge incorporation paradigm of SEAL~\citep{zweiger2025self}.
For each SQuAD context (document) $d$, the model first performs test-time weight updates on $d$ (knowledge incorporation),
and is then evaluated on its associated question set $\{q\}$ without providing the context in the prompt.
We report the mean answer correctness rate judged by GPT-4.1 as the evaluation metric.

\subsection{PaST Training Pipeline on Closed-Book SQuAD}
\label{app:squad:pipeline}

\newtcolorbox{promptbox}[1]{
  breakable,
  colback=gray!6,
  colframe=gray!55,
  title=\textbf{#1},
  fonttitle=\bfseries,
  left=1.2mm,right=1.2mm,top=1.0mm,bottom=1.0mm
}

\paragraph{Data used for skill distillation.}
To distill a domain-specific \emph{procedural} skill for parametric knowledge retrieval, we construct a source corpus
$\mathcal{D}^{\text{src}}$ consisting of $K=2$ rounds of SQuAD contexts, each with $N=50$ documents,
matching the data budget used in SEAL.
We denote the $k$-th round documents as $\mathcal{D}^{\text{src}}_k$.
All rounds use the same base model $\theta_{\text{base}}$ (Qwen2.5-7B) as the SFT initialization.

\paragraph{Two-round iterative refinement (PaST-$50\times 2$).}
Algorithm~\ref{alg:squad_past} summarizes the pipeline.
Each round performs: (i) knowledge injection via SFT on the documents, then (ii) skill acquisition via GRPO on the
corresponding closed-book QA pairs.
Crucially, the RL-induced parameter residual (skill vector) from the previous round is injected into the next round
\emph{after} SFT and \emph{before} RL, encouraging the learned skill to be content-invariant rather than overfitting to a single batch.

\begin{tcolorbox}[breakable, colback=gray!4, colframe=gray!60, title=\textbf{Notation (SQuAD)}]
\small
\begin{itemize}
    \item $\theta_{\text{base}}$: base model parameters.
    \item $\mathcal{D}^{\text{src}}_k$: source documents in round $k$ (each document is a SQuAD context).
    \item $\mathcal{Q}(\mathcal{D})$: the closed-book QA pairs attached to documents in $\mathcal{D}$.
    \item $\theta^{\text{sft}}_k$: SFT model trained on $\mathcal{D}^{\text{src}}_k$ (always initialized from $\theta_{\text{base}}$).
    \item $v_k$: skill vector after round $k$ (parameter residual capturing RL-induced procedural skill).
\end{itemize}
\end{tcolorbox}

\begin{algorithm}[t]
\caption{PaST iterative skill distillation on SQuAD (PaST-$50\times 2$).}
\label{alg:squad_past}
\small
\begin{enumerate}
    \item Initialize skill vector $v_0 \leftarrow 0$.
    \item For each round $k=1,\dots,K$ (here $K=2$):
    \begin{enumerate}
        \item \textbf{(Knowledge injection / SFT)} Train $\theta^{\text{sft}}_k \leftarrow \textsc{SFT}(\theta_{\text{base}}, \mathcal{D}^{\text{src}}_k)$.
        \item \textbf{(Skill carryover)} Initialize RL policy $\theta^{\text{init}}_k \leftarrow \theta^{\text{sft}}_k + v_{k-1}$.
        \item \textbf{(Skill acquisition / RL)} Train $\theta^{\text{rl}}_k \leftarrow \textsc{GRPO}(\theta^{\text{init}}_k, \mathcal{Q}(\mathcal{D}^{\text{src}}_k))$,
              where rewards are computed by GPT-4.1 judging answer correctness (Appendix~\ref{app:squad:prompts}).
        \item \textbf{(Skill extraction)} Update $v_k \leftarrow \theta^{\text{rl}}_k - \theta^{\text{sft}}_k$.
    \end{enumerate}
    \item Output final skill vector $v_{\star} \leftarrow v_K$.
\end{enumerate}
\end{algorithm}



\subsection{SFT Hyperparameters (Following SEAL)}
\label{app:squad:sft}

\paragraph{Synthetic data generation and packing.}
We generate implications for each passage and train on \textit{passage + implications}.
In the single-passage regime, we split the generated implication text by newlines into multiple
training sequences; in the multi-passage regime, we keep the full generation as one training document.
We sample $K=5$ synthetic generations per passage when forming the CPT training corpus.

\paragraph{Hyperparameters.}
We do not perform hyperparameter tuning on SQuAD.
For all SFT-based knowledge incorporation regimes, we directly adopt the best-performing hyperparameter configurations reported in SEAL.
Table~\ref{tab:squad_sft_hparams} summarizes the settings used in our implementation.

\paragraph{Compute and hardware.}
All experiments are run on NVIDIA A100 80GB GPUs.
For \textbf{single-passage} knowledge incorporation with LoRA, we use \textbf{two} A100 80GB GPUs:
one GPU hosts a \texttt{vLLM} inference server for fast generation, while the other GPU performs the inner-loop LoRA updates.
For \textbf{CPT} settings ($n{=}200/2067$), we run full fine-tuning on a single A100 80GB GPU.

\begin{table*}[t]
\centering
\caption{\textbf{Adopted SFT hyperparameters on SQuAD.} We do \emph{not} tune hyperparameters; instead, we directly reuse the best-performing configurations reported in SEAL for each regime.}
\label{tab:squad_sft_hparams}
\small
\begin{tabular}{lcc}
\toprule
\textbf{Hyperparameter} & \textbf{Single Passage ($n{=}1$, LoRA update)} & \textbf{CPT ($n{=}200/2067$, Full FT)} \\
\midrule
Base model & \multicolumn{2}{c}{\texttt{Qwen2.5-7B}} \\
Training data & \multicolumn{2}{c}{\textit{passage + synthetic implications} (same construction as SEAL)} \\
Max sequence length & 2048 & 2048 \\
\midrule
Update type & LoRA (test-time update) & Full fine-tuning \\
LoRA rank $r$ & 32 & -- \\
LoRA alpha $\alpha$ & 64 & -- \\
LoRA dropout & 0 & -- \\
LoRA target modules & as in SEAL implementation & -- \\
\midrule
Training epochs & 10  & 1 \\
Learning rate & $1\times 10^{-3}$ & $7\times 10^{-5}$ \\
Per-device batch size & 1 & 4 \\
Gradient accumulation & 1 & 2 \\
\bottomrule
\end{tabular}
\end{table*}

\subsection{GRPO (RL) Hyperparameters (Our Implementation)}
\label{app:squad:rl}

We implement GRPO training using \texttt{verl} with a custom reward function that queries an LLM judge
to score answer correctness. The effective hyperparameters are listed in Table~\ref{tab:squad_grpo_hparams} 

\begin{table*}[t]
\centering
\caption{GRPO hyperparameters used in our training script.}
\label{tab:squad_grpo_hparams}
\small
\begin{tabular}{ll}
\toprule
\textbf{Category} & \textbf{Value} \\
\midrule
Algorithm & GRPO (\texttt{algorithm.adv\_estimator=grpo}) \\
Epochs & 15 \\
Train batch size & 32 (\texttt{data.train\_batch\_size}) \\
Actor LR & $1\times 10^{-6}$ (\texttt{actor.optim.lr}) \\
Max prompt length & 512 (\texttt{data.max\_prompt\_length}) \\
Max response length & 1024 (\texttt{data.max\_response\_length}) \\
Rollout backend & vLLM (\texttt{actor\_rollout\_ref.rollout.name=vllm}) \\
\#rollouts per prompt ($n$) & 5 (\texttt{actor\_rollout\_ref.rollout.n}) \\
Sampling & temperature $=0.7$, top-$p=0.9$, top-$k=50$ \\
Max new tokens & 512 (\texttt{actor\_rollout\_ref.rollout.max\_new\_tokens}) \\
KL regularization & enabled (\texttt{use\_kl\_loss}), coef $=0.001$, type \texttt{low\_var\_kl} \\
PPO mini/micro batch & mini $=64$, micro per GPU $=8$ \\
Precision & rollout dtype \texttt{float16} \\
\bottomrule
\end{tabular}
\end{table*}

\subsection{Prompt Templates}
\label{app:squad:prompts}

\subsubsection{Implication generation prompt (for SFT data)}
\begin{promptbox}{Implications prompt}
\small
\begin{verbatim}
Let’s read the following passage and produce a 
list of implications derived directly or
indirectly from the content.
Passage:
{passage}
Implications:
\end{verbatim}
\end{promptbox}

\subsubsection{Closed-book QA prompt (actor rollout / evaluation)}
\begin{promptbox}{Closed-book QA prompt}
\small
\begin{verbatim}
Let’s answer a question directly and concisely.
Question: {question}
Answer:
\end{verbatim}
\end{promptbox}

\subsubsection{LLM-judge reward prompt (binary correctness)}
\begin{promptbox}{Judge prompt (returns yes/no)}
\small
\begin{verbatim}
You are a grading assistant. Your job is to 
determine whether a student’s answer correctly
answers the question based solely on the 
provided gold answer. Do not use any outside
knowledge. The student answer can include 
additional information, but it must at least 
fully convey the gold answer and must not 
contradict it. Ignore style, phrasing, or 
extra details that do not affect correctness. 
Respond ONLY with ‘yes’ or ‘no’.
Question: {question}
Gold answer: {gold}
Student answer: {pred}
Is the student answer correct based solely 
on the gold answer? Respond ‘yes’ or ‘no’.
\end{verbatim}
\end{promptbox}

%% file: src/appendix/loogle_settings.tex
\section{Implementation Details for LooGLE Experiments}
\label{app:loogle}

In this section, we provide comprehensive implementation details for our experiments on the LooGLE benchmark, including data selection, synthetic data generation pipelines, training hyperparameters, and prompt templates.

\subsection{Data Selection and Preprocessing}
\label{app:loogle_data}

\paragraph{Dataset Source.} 
We utilize the \textbf{Short Dependency QA} subset of the LooGLE benchmark \citep{li2024loogle}, which consists of long-context documents with an average length exceeding 21k tokens.

\paragraph{Data Splitting.}
To rigorously evaluate generalization, we implement a strict split:
\begin{itemize}
    \item \textbf{Source Domain (Training):} We select the \textbf{last 10 documents} (indices 100-104 for Round 1, indices 95-99 for Round 2) from the dataset. These documents are used solely for constructing the Skill Vector and are never seen during evaluation.
    \item \textbf{Target Domain (Evaluation):} We reserve the \textbf{first 50 documents} (indices 0-49) exclusively for testing.
\end{itemize}

\subsection{Synthetic Data Generation Pipeline}
\label{app:loogle_pipeline}

We employ a two-stage data generation strategy to create high-quality training signals for both SFT and RL. The training data for both stages is generated by the base model \verb|Qwen2.5-7B-Instruct| itself.

\paragraph{Stage 1: Multi-Task SFT Data Generation.}
To ensure the model deeply encodes the document content, we generate a diverse mixture of training tasks beyond simple text modeling. The SFT dataset $\mathcal{D}_{\text{SFT}}$ consists of three components mixed with specific ratios:
\begin{itemize}
    \item \textbf{Summarization (Ratio 50\%):} The model is tasked with compressing text chunks into concise summaries.
    \item \textbf{Recall/Expansion (Ratio 50\%):} The model is tasked with reconstructing detailed text from summaries (inverse of summarization).
\end{itemize}
This data is generated using a sliding window approach with chunk sizes of $\{1024, 2048, 4096\}$ tokens and an overlap of 256 tokens. 
We generate 16 data points for each chunk.
The generation temperature is set to $1.0$.

\paragraph{Stage 2: QA Generation for RL.}
For the RL stage, we generate synthetic Question-Answer pairs.
\begin{itemize}
    \item \textbf{Granularity:} We process the text in smaller chunks of $\{128, 256, 512\}$ tokens with an overlap of 16 tokens to capture fine-grained details.
    \item \textbf{Density:} We generate 8 pairs per chunk.
    \item \textbf{Prompt Diversity:} We employ a set of 6 distinct "Proposer" prompts (detailed in Sec.~\ref{app:loogle_prompts}) to ensure questions cover various aspects: factual details, reasoning (why/how), definitions, comparisons, lists, and significance.
\end{itemize}

\subsection{Prompt Templates}
\label{app:loogle_prompts}

To ensure the reproducibility of our synthetic data generation pipeline, we provide the exact content of the prompt templates used.

\paragraph{QA Generation Prompts.}
We utilize 6 distinct variations of prompts to generate diverse Question-Answer pairs. All variations share a common template to enforce strict formatting (XML tags) and specificity requirements. The full content of the instruction template and the 6 variations are detailed in Table~\ref{tab:qa_prompts_1} and \ref{tab:qa_prompts_2}.

\paragraph{Multi-Task SFT Prompts.}
For the summarization and expansion tasks in Stage 1 SFT, we randomly sample from a set of templates to prevent overfitting to a specific instruction format. The templates for Summarization and Recall/Expansion are listed in Table~\ref{tab:summary_recall_prompts}.

\input{tables/loogle_prompts}

\subsection{Training Hyperparameters}
\label{app:loogle_hyperparams}

We perform the training using 8 NVIDIA A100 GPUs. The Source Domain training (Stage 1 in our method) is further divided into two sub-phases to ensure stability:
\begin{enumerate}
    \item \textbf{SFT Phase 1 (Knowledge Encoding):} High learning rate training on the mixed dataset (Summarization, Recall, Verbatim) to enforce document memorization.
    \item \textbf{SFT Phase 2 (QA Adaptation):} Lower learning rate training specifically on the synthetic QA pairs to bridge the gap to the RL format.
    \item \textbf{RL Phase (Skill Sharpening):} GRPO training to refine the retrieval logic.
\end{enumerate}

\paragraph{Checkpoint Selection Strategy.}
To avoid overfitting to the source documents, we employ an independent validation set consisting of LooGLE documents with indices 90-94. We evaluate checkpoints every 40 steps. Based on the validation accuracy, we selected the checkpoint at \textbf{Step 120} for Round 1 and \textbf{Step 160} for Round 2 as the final models for skill extraction.

Table~\ref{tab:loogle_hyperparams_detailed} details the specific hyperparameters used in each phase.

\input{tables/loogle_hp}

\subsection{Evaluation}
\label{app:loogle_eval}

Given the open-ended nature of the generated answers in the LooGLE benchmark, standard metrics like Exact Match or ROUGE are often insufficient to capture semantic correctness. Therefore, we employ a Model-Based Evaluation paradigm using \verb|GPT-4.1| as an impartial judge to determine if the predicted answer matches the ground truth.

\paragraph{Judge Prompts.}
To ensure the evaluation is strictly binary and easy to parse, we enforce a strict output format. The exact prompts used for the judge model are provided below:

\begin{promptbox}{Evaluation Prompt}
\small
\begin{verbatim}
You are a precise evaluator. Your task is to 
deter-mine if the 'Predicted Answer' is 
semantically the same as the 'Ground Truth' for 
the given 'Question'. Your entire response MUST 
be only the single word 'True' or the single
word 'False'. Do not provide any explanation or
punctuation.
Question: \{question\}
Ground Truth: \{reference\}
Predicted Answer: \{pred\}
\end{verbatim}
\end{promptbox}

\paragraph{Robustness via Multi-Pass Sampling.}
To account for generation stochasticity and ensure the robustness of our reported metrics, we perform 3 independent generation runs with a temperature of $T=1.0$ for every question in the test set.This protocol ensures that our results reflect the model's consistent capability rather than lucky generations.

%% file: tables/loogle_prompts.tex
\begin{table*}[t!]
\centering
\begin{mybox}[Shared Instruction Template (Common to all QA Prompts)]{teal}
\begin{lstlisting}[style=trajectoryStyle]
You are an AI assistant tasked with generating a single, high-quality question-answer pair from a given text.

**Your instructions are critical:**
1.  **Generate one Q&A pair** based *only* on the provided text.
2.  **Specificity is Key:** The question *must* be self-contained and unambiguous. It must include specific names or key terms from the text (e.g., "What is 'Project Helios'?" instead of "What is this project?").
3.  **Format:** Your output *must* use XML-style tags: <question>Your question here</question> <answer>Your answer here</answer>. Do not include any other text before or after the tags.
4. [Specific Task Instruction: Refer to Variations 1-6 below]

**Example of a Good, Specific Q&A:**
[Task-specific Example: Refer to Variations 1-6 below]

**Text Fragment:**
{text_segment}

**Your Output:**
\end{lstlisting}
\end{mybox}

\vspace{-2mm}

\begin{mybox}[Variation 1: Factual Detail]{cyan}
\begin{lstlisting}[style=trajectoryStyle]
4.  **Task:** Focus on a specific, factual detail: a name, a date, a key term, or a specific component.

**Example of a Good, Specific Q&A:**
<question>
What consensus mechanism does the 'Helios' architecture pioneer?
</question>
<answer>
It pioneers a decentralized consensus mechanism called 'Proof-of-History' (PoH).
</answer>
\end{lstlisting}
\end{mybox}

\vspace{-2mm}

\begin{mybox}[Variation 2: Reasoning (Why/How)]{cyan}
\begin{lstlisting}[style=trajectoryStyle]
4.  **Task:** Focus on the *reason* (Why) or the *method* (How) behind a concept or event described in the text.

**Example of a Good, Specific Q&A:**
<question>
Why does the 'Proof-of-History' (PoH) mechanism successfully reduce latency?
</question>
<answer>
Because it creates a verifiable, sequential record of time, which avoids the need for solving complex computational puzzles like in 'Proof-of-Work'.
</answer>
\end{lstlisting}
\end{mybox}

\vspace{-4mm}
\caption{\textbf{QA Generation Prompts (Part 1).} The shared system instruction and the first two task variations used to generate synthetic QA pairs for LooGLE.}
\label{tab:qa_prompts_1}
\end{table*}

\begin{table*}[t]
\centering

\begin{mybox}[Variation 3: Definitions]{cyan}
\begin{lstlisting}[style=trajectoryStyle]
4.  **Task:** Generate a "What is..." or "What does... stand for?" question about a key concept.

**Example of a Good, Specific Q&A:**
<question>
What is 'Proof-of-History' (PoH) as described in the context of the 'Helios' architecture?
</question>
<answer>
It is a decentralized consensus mechanism that creates a verifiable, sequential record of time.
</answer>
\end{lstlisting}
\end{mybox}

\vspace{-2mm}

\begin{mybox}[Variation 4: Comparisons]{cyan}
\begin{lstlisting}[style=trajectoryStyle]
4.  **Task:** Focus on the *difference* or *similarity* between two specific concepts, methods, or entities in the text.

**Example of a Good, Specific Q&A:**
<question>
What is the key difference between the 'Proof-of-History' (PoH) mechanism and 'Proof-of-Work' (PoW)?
</question>
<answer>
'Proof-of-History' (PoH) creates a verifiable, sequential record of time, whereas 'Proof-of-Work' (PoW) relies on solving complex computational puzzles.
</answer>
\end{lstlisting}
\end{mybox}

\vspace{-2mm}

\begin{mybox}[Variation 5: Lists/Processes]{cyan}
\begin{lstlisting}[style=trajectoryStyle]
4.  **Task:** Generate a question that asks to list steps, components, or stages of a specific system or method.

**Example of a Good, Specific Q&A:**
<question>
What are the three main stages of the 'Project Nova' deployment pipeline?
</question>
<answer>
The three main stages are 'Build', 'Test', and 'Verify'.
</answer>
\end{lstlisting}
\end{mybox}

\vspace{-2mm}

\begin{mybox}[Variation 6: Significance/Impact]{cyan}
\begin{lstlisting}[style=trajectoryStyle]
4.  **Task:** Generate a "What is the significance of..." or "What is the main advantage of..." question.

**Example of a Good, Specific Q&A:**
<question>
What is the main advantage of the 'Helios' architecture's 'Proof-of-History' mechanism?
</question>
<answer>
Its main advantage is a drastic reduction in latency.
</answer>
\end{lstlisting}
\end{mybox}

\vspace{-4mm}
\caption{\textbf{QA Generation Prompts (Part 2).} Additional task variations (3-6) used to ensure diversity in the synthetic LooGLE training data.}
\label{tab:qa_prompts_2}
\end{table*}

\begin{table*}[t!]
\centering

\begin{mybox}[Summarization Templates]{orange}
\begin{lstlisting}[style=trajectoryStyle]
Create a concise and objective summary of the text below. Focus on the main ideas and most important information, presenting them clearly in your own words.\n\nText to summarize:\n{text_segment}\n\nNow generate your summary:

Distill the following text to its absolute essence. Provide a one-paragraph summary that captures the core argument and key takeaways. Avoid any fluff or secondary details.\n\nOriginal Text:\n{text_segment}\n\nYour distilled summary:

Break down the following text for someone completely unfamiliar with the topic. Your summary should be simple, clear, and use easy-to-understand language, as if you were explaining it to a 5th grader. Focus on the main concepts without getting lost in technical jargon.\n\nText:\n{text_segment}\n\nYour simple explanation:

Provide a summary of the following passage. Your summary should not only capture the key information but also reflect the original author's tone and style (e.g., formal, persuasive, humorous, critical). \n\nOriginal Passage:\n{text_segment}\n\nYour summary, in the original tone:

Provide a strictly neutral and objective summary of the provided text. Your goal is to recount the main points and arguments without injecting any personal opinion, interpretation, or evaluative language. Simply state what the text says.\n\nText:\n{text_segment}\n\nNeutral Summary:
\end{lstlisting}
\end{mybox}

\vspace{-2mm}

\begin{mybox}[Recall/Expansion Templates]{purple}
\begin{lstlisting}[style=trajectoryStyle]
Below is a summary of a larger text. Your task is to expand on this summary, creating a detailed and comprehensive paragraph that could have been the original source. Flesh out the main points with explanations, smooth transitions, and supporting details.\n\nSummary:\n{summary}\n\nNow generate the detailed original text:

You are given a concise summary below. Your objective is to generate a more detailed and fully-formed text based on it. Elaborate on the ideas presented in the summary, ensuring the resulting text is coherent, well-structured, and reads naturally.\n\nSummary:\n{summary}\n\nGenerated Text:

The following summary is a high-level overview of a piece of information. Your task is to reconstruct a plausible original text by filling in the necessary details, examples, and explanations that might have been removed during summarization.\n\nSummary:\n{summary}\n\nReconstructed Text with Details:

Consider the summary below as a set of conclusions or key statements. Your goal is to write a text that logically leads to these points. Build a coherent argument or explanation that culminates in the information provided in the summary.\n\nKey Points / Summary:\n{summary}\n\nText with Logical Development:

Take the core ideas presented in the summary below and weave them into a single, seamless paragraph. Focus on creating smooth transitions between the points so they flow together as a unified piece of writing, rather than a list of facts.\n\nCore Ideas:\n{summary}\n\nCoherent Paragraph:
\end{lstlisting}
\end{mybox}

\vspace{-4mm}
\caption{\textbf{Multi-Task SFT Prompts.} Templates used for the Summarization and Recall tasks in Stage 1 training to enhance knowledge encoding.}
\label{tab:summary_recall_prompts}
\end{table*}

%% file: tables/loogle_hp.tex
\begin{table*}[t!]
    \centering
    \small
    \caption{\textbf{Detailed Hyperparameters for LooGLE Experiments.} We employ a multi-stage curriculum: SFT Phase 1 enforces deep encoding of the 21k-token documents via mixed tasks; SFT Phase 2 adapts the model to the QA format; and the RL stage (GRPO) refines the retrieval logic using a closed-book setting (short prompt, long generation).}
    \label{tab:loogle_hyperparams_detailed}
    \begin{tabular}{lccc}
        \toprule
        \textbf{Hyperparameter} & \textbf{SFT Phase 1} & \textbf{SFT Phase 2} & \textbf{RL (GRPO)} \\
         & \textit{(Knowledge Encoding)} & \textit{(QA Adaptation)} & \textit{(Skill Sharpening)} \\
        \midrule
        \multicolumn{4}{l}{\textit{\textbf{General Configuration}}} \\
        Base Model & \multicolumn{3}{c}{Qwen2.5-7B-Instruct} \\
        Precision & bf16 & bf16 & bf16 \\
        Gradient Checkpointing & True & True & True \\
        Number of GPUs & 8 & 8 & 8 \\
        \midrule
        \multicolumn{4}{l}{\textit{\textbf{Optimization}}} \\
        Optimizer & AdamW & AdamW & AdamW \\
        Learning Rate & 1e-4 & 2e-5 & 1e-6 \\
        LR Scheduler & Cosine & Cosine & Constant \\
        Global Batch Size & 16 & 64 & 128 \\
        Micro Batch Size (per GPU) & 2 & 2 & 8 \\
        Total Epochs & 3 & 2 & 10 (Selected Best Step) \\
        \midrule
        \multicolumn{4}{l}{\textit{\textbf{Data \& Context}}} \\
        Max Sequence Length & 8192 & 8192 & - \\
        Max Prompt Length & - & - & 128 (Closed-Book) \\
        Max Response Length & - & - & 1024 \\
        Data Source & Mixed (Sum/Recall) & Synthetic QA & Synthetic QA \\
        \midrule
        \multicolumn{4}{l}{\textit{\textbf{RL Specifics (GRPO)}} } \\
        Group Size ($G$) & - & - & 5 \\
        KL Coefficient ($\beta$) & - & - & 0.001 \\
        KL Reference Model & - & - & SFT Phase 2 Checkpoint \\
        Reward Function & - & - & GPT-4.1 Judge \\
        \bottomrule
    \end{tabular}
\end{table*}

%% file: src/appendix/toolbench_settings.tex
\section{ToolBench Experiments}
\label{app:toolbench}

\subsection{Category Filtering and Selection Criteria}
To ensure a robust and balanced evaluation of cross-domain generalization in agentic tool use, we perform a multi-stage filtering process on the original ToolBench and StableToolBench datasets. We apply the following constraints to select the evaluation categories:

\begin{itemize}
    \item \textbf{Data Sufficiency:} We exclude categories with fewer than 3 solvable queries in the StableToolBench test set to ensure that the evaluation results are statistically meaningful.
    \item \textbf{Category Complexity:} We select categories with an API count between 75 and 350. This range ensures that the domain is sufficiently complex to require parametric internalizing (avoiding trivial domains with too few APIs) while remaining manageable for the environment simulator.
\end{itemize}

Based on these criteria, 21 categories were retained. We designate \textit{Movies} as the \textbf{Source Domain} for skill acquisition because its API count (111) represents the median complexity of the filtered set, and it contains a relatively high number of test queries (35), allowing for stable monitoring of the RL training progress. The remaining 20 categories serve as the \textbf{Target Domains} for zero-shot generalization testing.

\paragraph{Statistics of Evaluated Categories}
Table~\ref{tab:category_stats} summarizes the statistics for the selected domains. The "APIs" column denotes the number of unique API schemas the model must internalize, and the "Test Queries" column denotes the number of solvable scenarios used for final evaluation.

\input{tables/toolbench_categories}

\subsection{Implementation Details for SFT}
\label{app:sft_details}

Our SFT process is divided into two stages: \textbf{Knowledge Internalization} (Stage 1) and \textbf{Format Alignment} (Stage 2). All experiments are conducted using the \texttt{verl} framework with FSDP2 strategy on 8$\times$A100 (80GB) GPUs.

\subsubsection{Data Generation for Stage 1}
To internalize the parametric knowledge of tools, we use the base model \verb|Qwen2.5-7B-Instruct| to transform raw JSON schemas into diverse natural language (NL) descriptions and QA pairs.

\paragraph{Teacher Prompts.} 
We use three distinct templates to generate descriptions from JSON schemas to ensure linguistic diversity.

\begin{promptbox}{Teacher Generation Prompts}
\small
Transform the following API JSON into a coherent, natural language paragraph describing how to use it. CRITICAL: You MUST explicitly mention the API name (``\{API\_NAME\}'') and its purpose at the beginning. \\

You are a technical documentation expert. Convert the provided API JSON definition into a comprehensive technical reference. Rules: 1. Identity: Explicitly state the API Name. 2. Purpose: Explain functionality. 3. Inputs: Detail parameters. \\

Convert the provided API definition into a structured natural language summary. Format: 1. Tool Identifier: The exact API name string. 2. Intent: User goal. 3. Action: Parameters.
\end{promptbox}

\paragraph{Training Pair Construction.}
Based on the generated descriptions, we construct four types of training pairs to build a robust mapping between API names, intents, and schemas:
\begin{itemize}
    \item \textbf{Type A (Name $\to$ Usage):} Queries like "How do I use the \{name\} API?" mapped to NL descriptions.
    \item \textbf{Type B (Intent $\to$ Usage):} Queries like "Identify the API defined by: \{description\} and explain its parameters."
    \item \textbf{Type C (Intent $\to$ Raw JSON):} Requesting the underlying JSON schema based on a description.
    \item \textbf{Type D (Name $\to$ Raw JSON):} Directly mapping the API name to its original JSON definition.
\end{itemize}

\subsubsection{Training Configurations and Hyperparameters}
We utilize different optimization strategies for the two stages. Stage 1 focuses on broad knowledge acquisition with a larger batch size and higher learning rate, while Stage 2 performs fine-grained format alignment.
The detailed configurations are shown in Table~\ref{tab:toolbench_sft_hp}.

\input{tables/toolbench_sft_hp}

\subsection{Details of Reinforcement Learning for Tool Use}
\label{app:ppo_details}

We implement the reinforcement learning phase based on the Search-R1~\citep{jin2025search} framework, extending it to support multi-turn agentic tool-use trajectories and environment interactions.

\paragraph{Agent Prompting.} 
The model is prompted to rely on its internalized knowledge acquired during the SFT phase. The system prompt specifies the ReAct format (\texttt{Thought}, \texttt{Action}, \texttt{Action Input}) and lists only the names of available APIs. The exact prompt template is provided in Table~\ref{tab:agent_prompt}.

\input{tables/agent_prompt}

\paragraph{Environment Simulator.}
Since real-world API execution can be unstable or costly during RL, we employ \texttt{gpt-4o-mini} as an API Simulator. The simulator is provided with the full API documentation and examples (from the original ToolBench) and is tasked to validate the agent's generated \texttt{Action Input} against the ground-truth schema. It returns either a realistic JSON response or a specific error message (e.g., missing required parameters, type mismatch). The simulator's instructions are detailed in Table~\ref{tab:api_simulator_prompt}.

\input{tables/api_simulator_prompt}

\subsubsection{Reward Design}
The reward signal $R$ for a trajectory is composed of three components, designed to encourage format adherence, successful tool invocation, and task resolution.

\paragraph{Format and Execution Reward.}
For each intermediate turn $j$, we assign a step-wise reward $R_{\text{step}, j}$ based on the validity of the generated action and its execution outcome. Specifically:
(i) a positive reward of \textbf{+0.1} is granted if the action follows the correct ReAct format and the API call is successfully executed by the simulator; 
(ii) a penalty of \textbf{$-0.1$} is applied if the format is correct but the API call fails (e.g., due to missing required parameters or type mismatches); 
(iii) a heavier penalty of \textbf{$-0.2$} is imposed if the model fails to follow the output format or invokes a hallucinated API name not present in the available toolset.

\paragraph{Termination Reward.}
To prevent infinite loops and encourage concise solutions, we apply a maximum of 5 turns and apply a reward at the final turn where the model calls the \texttt{Finish} tool:
\begin{itemize}
    \item \textbf{Active Finish:} If the model successfully calls \texttt{Finish}, it receives $+0.2$.
    \item \textbf{Forced Termination:} If the model reaches the maximum turn limit without calling \texttt{Finish}, it receives $-0.5$.
\end{itemize}

\paragraph{Solution Reward.}
The final Pass Reward is determined by a GPT-4.1 judge, which evaluates the entire trajectory. The judge assigns one of three statuses based on the rules in Table~\ref{tab:toolbench_eval_prompt}: \texttt{Solved} ($+1.0$), \texttt{Partially Solved} ($+0.5$), or \texttt{Unsolved} ($0.0$).

\input{tables/toolbench_eval_prompt}

\subsection{Hyperparameters for Reinforcement Learning}
\label{subsec:rl_hyperparams}

The reinforcement learning phase is conducted using the \methodShort framework on 4$\times$A100 GPUs. Table~\ref{tab:toolbench_rl_hp} summarizes the detailed hyperparameter settings used for the ToolBench experiments.

\input{tables/toolbench_rl_hp}

%% file: tables/toolbench_categories.tex
\begin{table*}[ht]
\centering
\caption{Statistics of the 21 selected categories from ToolBench. The Source Domain is used for RL training, while Target Domains are used for zero-shot evaluation.}
\begin{tabular}{lccc}
\toprule
\textbf{Category} & \textbf{Domain Role} & \textbf{\# APIs} & \textbf{\# Test Queries} \\
\midrule
Movies & \textbf{Source} & 111 & 35 \\
\midrule
Advertising & Target & 118 & 3 \\
AI \& ML & Target & 108 & 4 \\
Business Software & Target & 199 & 5 \\
Communication & Target & 250 & 8 \\
Database & Target & 260 & 7 \\
Education & Target & 214 & 9 \\
Email & Target & 143 & 5 \\
Financial & Target & 224 & 11 \\
Food & Target & 208 & 9 \\
Health and Fitness & Target & 90 & 7 \\
Location & Target & 328 & 11 \\
Mapping & Target & 113 & 7 \\
Media & Target & 159 & 12 \\
SMS & Target & 75 & 3 \\
Science & Target & 99 & 4 \\
Search & Target & 103 & 13 \\
Text Analysis & Target & 98 & 4 \\
Video Images & Target & 207 & 44 \\
Weather & Target & 199 & 8 \\
eCommerce & Target & 342 & 15 \\
\bottomrule
\end{tabular}
\label{tab:category_stats}
\end{table*}

%% file: tables/toolbench_sft_hp.tex
\begin{table*}[ht]
\centering
\small
\caption{Hyperparameters for the two stages of SFT in ToolBench.}
\begin{tabular}{lcc}
\toprule
\textbf{Hyperparameter} & \textbf{Stage 1: Internalization} & \textbf{Stage 2: Alignment} \\
\midrule
Base Model & \multicolumn{2}{c}{Qwen2.5-7B-Instruct} \\
Max Sequence Length & 8192 & 8192 \\
Optimizer & \multicolumn{2}{c}{AdamW} \\
Precision & \multicolumn{2}{c}{BF16} \\
Learning Rate & $5 \times 10^{-5}$ & $2 \times 10^{-5}$ \\
Total Batch Size & 64 & 32 \\
Micro Batch Size (per GPU) & 2 & 2 \\
Total Epochs & 3 & 3 \\
Parallel Strategy & \multicolumn{2}{c}{FSDP2 (verl)} \\
Liger Kernel & \multicolumn{2}{c}{Enabled} \\
\bottomrule
\end{tabular}
\label{tab:toolbench_sft_hp}
\end{table*}

%% file: tables/agent_prompt.tex
\begin{table*}[t!]
\centering
\caption{The system prompt for the agent in ToolBench Experiments.}
\begin{mybox}[Agent System Prompt]{teal}
\begin{lstlisting}[style=trajectoryStyle]
You are an intelligent agent designed to handle real-time user queries using a variety of tools.

First, you will receive a task description. Then, you will enter a loop of reasoning and acting to complete the task.

At each step, follow this process:
1. **Thought**: Analyze the current status and determine the next logical step.
2. **Action**: Select the appropriate tool to execute that step and output the function name directly.
3. **Action Input**: Provide the arguments for the tool as a STRICT valid JSON object.

Output Format:
Thought: <your reasoning>
Action: <function_name>
Action Input: <function_arguments_as_a_valid_JSON_object>

After the action is executed, you will receive the result ('Observation: <observation>'). Based on the new state, continue the loop until the task is complete.

Constraints & Rules:
1. **Action Field**: The "Action" output must be the EXACT name of the function. Do NOT include parentheses `()`, words like "call" or "use", or any punctuation.
2. **Finishing**: You MUST call the "Finish" function to submit your final answer.

Available Tools:
1. **General Tools**: You have been trained on a specific set of APIs: {api_names}. You must rely on your **internal knowledge** to recall the correct parameter schemas for these tools.
2. **Termination Tool**: You MUST use the following tool to finish the task. Its definition is provided below:
{"name": "Finish", "description": "If you believe that you have obtained a result that can answer the task, please call this function to provide the final answer. Remember: you must ALWAYS call this function at the end of your attempt, and the only part that will be shown to the user is the final answer, so it should contain sufficient information.", "parameters": {"properties": {"final_answer": {"type": "string", "description": "The final answer you want to give the user."}}}, "required": ["final_answer"], "optional": []}
\end{lstlisting}
\end{mybox}
\vspace{-4mm}
\label{tab:agent_prompt}
\end{table*}

%% file: tables/api_simulator_prompt.tex
\begin{table*}[t!]
\centering
\caption{The simulator's instructions in ToolBench Experiments.}
\begin{mybox}[API Simulator Instructions]{cyan}
\begin{lstlisting}[style=trajectoryStyle]
You are an advanced API Simulator and Validator. Your role is to act as a real API server, strictly adhering to the provided documentation to process requests.

### 1. Input Structure Explanation
The user will provide input in the following specific format:
API Documentation:
Contains the API's URL, method, description, and parameter definitions (required/optional).
API Examples:
Contains reference calls (Note: This section is truncated to 2048 chars and may be incomplete; use it for style reference but rely on Documentation for logic).
API Input:
The specific arguments/payload you need to process.

### 2. Processing Logic
1. **Analyze:** Read the `API Documentation` to understand the schema and constraints (types, required fields).
2. **Validate:** Check the `API Input` against the `API Documentation`.
- Are all `required_parameters` present?
- Do the data types match (e.g., string vs int)?
3. **Execute:**
- **If Valid:** Generate a realistic, rich JSON response. 
- **If Invalid:** Generate a JSON response where `error` describes the specific validation failure.

### 3. Output Format
You must output ONLY a valid JSON object. No Markdown code blocks. No conversational text.

**JSON Schema:**
{
    "error": "String describing the error (if any), otherwise empty string",
    "response": <The_Simulated_Response_Object_or_Null>
}

### 4. Behavior Rules
- **Length Constraints:** - Keep the response **concise and lightweight**. Keep the entire JSON output shorter than 300 words. Do not generate excessively large payloads. If the API returns a list or array, **limit it to maximum 1-2 items**.
- **Source of Truth:** Do not blindly copy "API Examples" if they contradict the "API Input". The "API Input" is your priority.
\end{lstlisting}
\end{mybox}

\vspace{-4mm}
\label{tab:api_simulator_prompt}
\end{table*}

%% file: tables/toolbench_eval_prompt.tex
\begin{table*}[t!]
\centering
\caption{The evaluation prompt in ToolBench experiments.}
\begin{mybox}[GPT-4.1 Judge Prompt]{cyan}
\begin{lstlisting}[style=trajectoryStyle]
Giving the query and the corresponding execution trajectory (including thoughts, tool calls, and observations), evaluate the `answer_status` based on these rules:

1. **Solved**: The tool calls were successful. The final answer is strictly grounded in the real "Observation" data and fully addresses the query.
2. **Partially Solved**: The model used real "Observation" data, but the task is only halfway finished or the final answer missed some details from the observations.
3. **Unsolved**: 
    - The model fabricated information not found in the Observations.
    - The tool calls failed, and the model failed to solve the query or made up a result.
    - The answer is incorrect or irrelevant.

Output a JSON object with the following fields:
- "reason": A very brief explanation (less than 20 words).
- "answer_status": One of ["Solved", "Partially Solved", "Unsolved"].

<Target_Query>
{query}
</Target_Query>

<Model_Execution_Trajectory>
{trajectory}
</Model_Execution_Trajectory>
\end{lstlisting}
\end{mybox}
\vspace{-4mm}
\label{tab:toolbench_eval_prompt}
\end{table*}

%% file: tables/toolbench_rl_hp.tex
\begin{table*}[ht]
\centering
\small
\caption{Detailed hyperparameters for PPO training in ToolBench experiments. }
\begin{tabular}{llc}
\toprule
\textbf{Category} & \textbf{Hyperparameter} & \textbf{Value} \\
\midrule
\multirow{4}{*}{Data \& Environment} & Max Prompt Length & 8192 \\
 & Max Response Length & 1024 \\
 & \textbf{Max Simulation Turns} & 5 \\
\midrule
\multirow{6}{*}{Optimization} & Actor Learning Rate & $1 \times 10^{-6}$ \\
 & Critic Learning Rate & $1 \times 10^{-5}$ \\
 & Actor LR Warmup Ratio & 0.285 \\
 & Critic LR Warmup Ratio & 0.015 \\
 & LR Scheduler & Constant \\
 & Optimizer & AdamW \\
\midrule
\multirow{6}{*}{PPO Algorithm} & Advantage Estimator & GAE \\
 & PPO Epochs per Batch & 1 \\
 & Mini-batch Size & 64 \\
 & Clip Range ($\epsilon$) & 0.2 \\
 & KL Penalty Coefficient ($\beta$) & 0.001 \\
 & Entropy Coefficient & 0.001 \\
\midrule
\multirow{2}{*}{Training Schedule} & Total Training Steps & 180 \\
 & Compute Precision & BF16 \\
\bottomrule
\end{tabular}
\label{tab:toolbench_rl_hp}
\end{table*}

%% file: src/appendix/ai_usage.tex
\section{Declaration of AI Usage}

Generative AI tools were used for grammar refinement and language polishing to enhance the readability of the manuscript. AI assistance was also employed during the coding and implementation phases of the project. All AI-assisted outputs were reviewed by the authors to ensure the technical quality and accuracy of the final paper.

%% file: custom.bib
@article{ilharco2022editing,
  title={Editing models with task arithmetic},
  author={Ilharco, Gabriel and Ribeiro, Marco Tulio and Wortsman, Mitchell and Gururangan, Suchin and Schmidt, Ludwig and Hajishirzi, Hannaneh and Farhadi, Ali},
  journal={arXiv preprint arXiv:2212.04089},
  year={2022}
}

@article{du2025knowledge,
  title={Knowledge grafting of large language models},
  author={Du, Guodong and Zhou, Xuanning and Li, Junlin and Li, Zhuo and Shi, Zesheng and Lin, Wanyu and Tang, Ho-Kin and Li, Xiucheng and Liu, Fangming and Wang, Wenya and others},
  journal={arXiv preprint arXiv:2505.18502},
  year={2025}
}

@inproceedings{cao2025param,
  title={Param $\delta$ for Direct Mixing: Post-Train Large Language Model At Zero Cost},
  author={Cao, Sheng and Wu, Mingrui and Prasad, Karthik and Tian, Yuandong and Liu, Zechun},
  booktitle={The Thirteenth International Conference on Learning Representations},
  year={2025}
}

@article{zbeeb2025reasoning,
  title={Reasoning Vectors: Transferring Chain-of-Thought Capabilities via Task Arithmetic},
  author={Zbeeb, Mohammad and Hammoud, Hasan Abed Al Kader and Ghanem, Bernard},
  journal={arXiv preprint arXiv:2509.01363},
  year={2025}
}

@article{zhu2025path,
  title={The path not taken: Rlvr provably learns off the principals},
  author={Zhu, Hanqing and Zhang, Zhenyu and Huang, Hanxian and Su, DiJia and Liu, Zechun and Zhao, Jiawei and Fedorov, Igor and Pirsiavash, Hamed and Sha, Zhizhou and Lee, Jinwon and others},
  journal={arXiv preprint arXiv:2511.08567},
  year={2025}
}

@article{cheng2023adapting,
  title={Adapting large language models to domains via reading comprehension},
  author={Cheng, Daixuan and Huang, Shaohan and Wei, Furu},
  journal={arXiv preprint arXiv:2309.09530},
  year={2023}
}

@article{meng2022locating,
  title={Locating and editing factual associations in gpt},
  author={Meng, Kevin and Bau, David and Andonian, Alex and Belinkov, Yonatan},
  journal={Advances in neural information processing systems},
  volume={35},
  pages={17359--17372},
  year={2022}
}

@inproceedings{
meng2023massediting,
title={Mass-Editing Memory in a Transformer},
author={Kevin Meng and Arnab Sen Sharma and Alex J Andonian and Yonatan Belinkov and David Bau},
booktitle={The Eleventh International Conference on Learning Representations },
year={2023},
url={https://openreview.net/forum?id=MkbcAHIYgyS}
}

@inproceedings{yehudai2024achieving,
  title={Achieving human parity in content-grounded datasets generation},
  author={Yehudai, Asaf and Carmeli, Boaz and Mass, Yosi and Arviv, Ofir and Mills, Nathaniel and Shnarch, Eyal and Choshen, Leshem},
  booktitle={International Conference on Learning Representations},
  year={2024}
}

@article{lampinen2025generalization,
  title={On the generalization of language models from in-context learning and finetuning: a controlled study},
  author={Lampinen, Andrew K and Chaudhry, Arslan and Chan, Stephanie CY and Wild, Cody and Wan, Diane and Ku, Alex and Bornschein, J{\"o}rg and Pascanu, Razvan and Shanahan, Murray and McClelland, James L},
  journal={arXiv preprint arXiv:2505.00661},
  year={2025}
}

@article{zweiger2025self,
  title={Self-Adapting Language Models},
  author={Zweiger, Adam and Pari, Jyothish and Guo, Han and Aky{\"u}rek, Ekin and Kim, Yoon and Agrawal, Pulkit},
  journal={arXiv preprint arXiv:2506.10943},
  year={2025}
}

@article{ouyang2022training,
  title={Training language models to follow instructions with human feedback},
  author={Ouyang, Long and Wu, Jeff and Jiang, Xu and Almeida, Diogo and Wainwright, Carroll L and Mishkin, Pamela and Zhang, Chong and Agarwal, Sandhini and Slama, Katarina and Ray, Alex and others},
  journal={Advances in Neural Information Processing Systems},
  volume={35},
  pages={27730--27744},
  year={2022}
}

@article{touvron2023llama,
  title={Llama: Open and efficient foundation language models},
  author={Touvron, Hugo and Lavril, Thibaut and Izacard, Gautier and Martinet, Xavier and Lachaux, Marie-Anne and Lacroix, Timoth{\'e}e and Rozi{\`e}re, Baptiste and Goyal, Naman and Hambro, Eric and Azhar, Faisal and others},
  journal={arXiv preprint arXiv:2302.13971},
  year={2023}
}

@article{lewis2020retrieval,
  title={Retrieval-augmented generation for knowledge-intensive nlp tasks},
  author={Lewis, Patrick and Perez, Ethan and Piktus, Aleksandra and Petroni, Fabio and Karpukhin, Vladimir and Goyal, Naman and K{\"u}ttler, Heinrich and Lewis, Mike and Yih, Wen-tau and Rockt{\"a}schel, Tim and others},
  journal={Advances in neural information processing systems},
  volume={33},
  pages={9459--9474},
  year={2020}
}

@article{shao2023enhancing,
  title={Enhancing retrieval-augmented large language models with iterative retrieval-generation synergy},
  author={Shao, Zhihong and Gong, Yeyun and Shen, Yelong and Huang, Minlie and Duan, Nan and Chen, Weizhu},
  journal={arXiv preprint arXiv:2305.15294},
  year={2023}
}

@article{yao2023editing,
  title={Editing large language models: Problems, methods, and opportunities},
  author={Yao, Yunzhi and Wang, Peng and Tian, Bozhong and Cheng, Siyuan and Li, Zhoubo and Deng, Shumin and Chen, Huajun and Zhang, Ningyu},
  journal={arXiv preprint arXiv:2305.13172},
  year={2023}
}

@article{mao2025lift,
  title={Lift: Improving long context understanding of large language models through long input fine-tuning},
  author={Mao, Yansheng and Xu, Yufei and Li, Jiaqi and Meng, Fanxu and Yang, Haotong and Zheng, Zilong and Wang, Xiyuan and Zhang, Muhan},
  journal={arXiv preprint arXiv:2502.14644},
  year={2025}
}

@article{chu2025sft,
  title={Sft memorizes, rl generalizes: A comparative study of foundation model post-training},
  author={Chu, Tianzhe and Zhai, Yuexiang and Yang, Jihan and Tong, Shengbang and Xie, Saining and Schuurmans, Dale and Le, Quoc V and Levine, Sergey and Ma, Yi},
  journal={arXiv preprint arXiv:2501.17161},
  year={2025}
}

@article{schulman2017proximal,
  title={Proximal policy optimization algorithms},
  author={Schulman, John and Wolski, Filip and Dhariwal, Prafulla and Radford, Alec and Klimov, Oleg},
  journal={arXiv preprint arXiv:1707.06347},
  year={2017}
}

@article{vaswani2017attention,
  title={Attention is all you need},
  author={Vaswani, Ashish and Shazeer, Noam and Parmar, Niki and Uszkoreit, Jakob and Jones, Llion and Gomez, Aidan N and Kaiser, {\L}ukasz and Polosukhin, Illia},
  journal={Advances in neural information processing systems},
  volume={30},
  year={2017}
}

@article{cheng2024dated,
  title={Dated data: Tracing knowledge cutoffs in large language models},
  author={Cheng, Jeffrey and Marone, Marc and Weller, Orion and Lawrie, Dawn and Khashabi, Daniel and Van Durme, Benjamin},
  journal={arXiv preprint arXiv:2403.12958},
  year={2024}
}

@article{brown2020language,
  title={Language models are few-shot learners},
  author={Brown, Tom and Mann, Benjamin and Ryder, Nick and Subbiah, Melanie and Kaplan, Jared D and Dhariwal, Prafulla and Neelakantan, Arvind and Shyam, Pranav and Sastry, Girish and Askell, Amanda and others},
  journal={Advances in neural information processing systems},
  volume={33},
  pages={1877--1901},
  year={2020}
}

@article{shao2024deepseekmath,
  title={Deepseekmath: Pushing the limits of mathematical reasoning in open language models},
  author={Shao, Zhihong and Wang, Peiyi and Zhu, Qihao and Xu, Runxin and Song, Junxiao and Bi, Xiao and Zhang, Haowei and Zhang, Mingchuan and Li, YK and Wu, Yang and others},
  journal={arXiv preprint arXiv:2402.03300},
  year={2024}
}

@article{guo2025deepseek,
  title={Deepseek-r1: Incentivizing reasoning capability in llms via reinforcement learning},
  author={Guo, Daya and Yang, Dejian and Zhang, Haowei and Song, Junxiao and Zhang, Ruoyu and Xu, Runxin and Zhu, Qihao and Ma, Shirong and Wang, Peiyi and Bi, Xiao and others},
  journal={arXiv preprint arXiv:2501.12948},
  year={2025}
}

@article{wei2025webagent,
  title={Webagent-r1: Training web agents via end-to-end multi-turn reinforcement learning},
  author={Wei, Zhepei and Yao, Wenlin and Liu, Yao and Zhang, Weizhi and Lu, Qin and Qiu, Liang and Yu, Changlong and Xu, Puyang and Zhang, Chao and Yin, Bing and others},
  journal={arXiv preprint arXiv:2505.16421},
  year={2025}
}

@article{qian2025toolrl,
  title={Toolrl: Reward is all tool learning needs},
  author={Qian, Cheng and Acikgoz, Emre Can and He, Qi and Wang, Hongru and Chen, Xiusi and Hakkani-T{\"u}r, Dilek and Tur, Gokhan and Ji, Heng},
  journal={arXiv preprint arXiv:2504.13958},
  year={2025}
}

@inproceedings{
mukherjee2025reinforcement,
title={Reinforcement Learning Finetunes Small Subnetworks in Large Language Models},
author={Sagnik Mukherjee and Lifan Yuan and Dilek Hakkani-T{\"u}r and Hao Peng},
booktitle={The Thirty-ninth Annual Conference on Neural Information Processing Systems},
year={2025},
url={https://openreview.net/forum?id=0NdS4xCngO}
}

@article{shenfeld2025rl,
  title={RL's Razor: Why Online Reinforcement Learning Forgets Less},
  author={Shenfeld, Idan and Pari, Jyothish and Agrawal, Pulkit},
  journal={arXiv preprint arXiv:2509.04259},
  year={2025}
}

@article{rajpurkar2016squad,
  title={Squad: 100,000+ questions for machine comprehension of text},
  author={Rajpurkar, Pranav and Zhang, Jian and Lopyrev, Konstantin and Liang, Percy},
  journal={arXiv preprint arXiv:1606.05250},
  year={2016}
}

@inproceedings{li2024loogle,
  title={Loogle: Can long-context language models understand long contexts?},
  author={Li, Jiaqi and Wang, Mengmeng and Zheng, Zilong and Zhang, Muhan},
  booktitle={Proceedings of the 62nd Annual Meeting of the Association for Computational Linguistics (Volume 1: Long Papers)},
  pages={16304--16333},
  year={2024}
}

@article{guo2024stabletoolbench,
  title={Stabletoolbench: Towards stable large-scale benchmarking on tool learning of large language models},
  author={Guo, Zhicheng and Cheng, Sijie and Wang, Hao and Liang, Shihao and Qin, Yujia and Li, Peng and Liu, Zhiyuan and Sun, Maosong and Liu, Yang},
  journal={arXiv preprint arXiv:2403.07714},
  year={2024}
}

@article{qin2023toolllm,
  title={Toolllm: Facilitating large language models to master 16000+ real-world apis},
  author={Qin, Yujia and Liang, Shihao and Ye, Yining and Zhu, Kunlun and Yan, Lan and Lu, Yaxi and Lin, Yankai and Cong, Xin and Tang, Xiangru and Qian, Bill and others},
  journal={arXiv preprint arXiv:2307.16789},
  year={2023}
}

@article{liu2021ttt++,
  title={Ttt++: When does self-supervised test-time training fail or thrive?},
  author={Liu, Yuejiang and Kothari, Parth and Van Delft, Bastien and Bellot-Gurlet, Baptiste and Mordan, Taylor and Alahi, Alexandre},
  journal={Advances in Neural Information Processing Systems},
  volume={34},
  pages={21808--21820},
  year={2021}
}

@inproceedings{osowiechi2023tttflow,
  title={Tttflow: Unsupervised test-time training with normalizing flow},
  author={Osowiechi, David and Hakim, Gustavo A Vargas and Noori, Mehrdad and Cheraghalikhani, Milad and Ben Ayed, Ismail and Desrosiers, Christian},
  booktitle={Proceedings of the IEEE/CVF Winter Conference on Applications of Computer Vision},
  pages={2126--2134},
  year={2023}
}

@article{gandelsman2022test,
  title={Test-time training with masked autoencoders},
  author={Gandelsman, Yossi and Sun, Yu and Chen, Xinlei and Efros, Alexei},
  journal={Advances in Neural Information Processing Systems},
  volume={35},
  pages={29374--29385},
  year={2022}
}

@inproceedings{hong2023mecta,
  title={Mecta: Memory-economic continual test-time model adaptation},
  author={Hong, Junyuan and Lyu, Lingjuan and Zhou, Jiayu and Spranger, Michael},
  booktitle={2023 International Conference on Learning Representations},
  year={2023}
}

@article{roberts2020much,
  title={How much knowledge can you pack into the parameters of a language model?},
  author={Roberts, Adam and Raffel, Colin and Shazeer, Noam},
  journal={arXiv preprint arXiv:2002.08910},
  year={2020}
}

@article{schick2023toolformer,
  title={Toolformer: Language models can teach themselves to use tools},
  author={Schick, Timo and Dwivedi-Yu, Jane and Dess{\`\i}, Roberto and Raileanu, Roberta and Lomeli, Maria and Hambro, Eric and Zettlemoyer, Luke and Cancedda, Nicola and Scialom, Thomas},
  journal={Advances in Neural Information Processing Systems},
  volume={36},
  pages={68539--68551},
  year={2023}
}

@article{li2023api,
  title={Api-bank: A comprehensive benchmark for tool-augmented llms},
  author={Li, Minghao and Zhao, Yingxiu and Yu, Bowen and Song, Feifan and Li, Hangyu and Yu, Haiyang and Li, Zhoujun and Huang, Fei and Li, Yongbin},
  journal={arXiv preprint arXiv:2304.08244},
  year={2023}
}

@book{golub2013matrix,
  title={Matrix computations},
  author={Golub, Gene H and Van Loan, Charles F},
  year={2013},
  publisher={JHU press}
}

@article{ba2016layer,
  title={Layer normalization},
  author={Ba, Jimmy Lei and Kiros, Jamie Ryan and Hinton, Geoffrey E},
  journal={arXiv preprint arXiv:1607.06450},
  year={2016}
}

@book{vershynin2018high,
  title={High-dimensional probability: An introduction with applications in data science},
  author={Vershynin, Roman},
  volume={47},
  year={2018},
  publisher={Cambridge university press}
}

@misc{qwen2025qwen25technicalreport,
      title={Qwen2.5 Technical Report}, 
      author={Qwen and : and An Yang and Baosong Yang and Beichen Zhang and Binyuan Hui and Bo Zheng and Bowen Yu and Chengyuan Li and Dayiheng Liu and Fei Huang and Haoran Wei and Huan Lin and Jian Yang and Jianhong Tu and Jianwei Zhang and Jianxin Yang and Jiaxi Yang and Jingren Zhou and Junyang Lin and Kai Dang and Keming Lu and Keqin Bao and Kexin Yang and Le Yu and Mei Li and Mingfeng Xue and Pei Zhang and Qin Zhu and Rui Men and Runji Lin and Tianhao Li and Tianyi Tang and Tingyu Xia and Xingzhang Ren and Xuancheng Ren and Yang Fan and Yang Su and Yichang Zhang and Yu Wan and Yuqiong Liu and Zeyu Cui and Zhenru Zhang and Zihan Qiu},
      year={2025},
      eprint={2412.15115},
      archivePrefix={arXiv},
      primaryClass={cs.CL},
      url={https://arxiv.org/abs/2412.15115}, 
}

@article{jin2025search,
  title={Search-r1: Training llms to reason and leverage search engines with reinforcement learning},
  author={Jin, Bowen and Zeng, Hansi and Yue, Zhenrui and Yoon, Jinsung and Arik, Sercan and Wang, Dong and Zamani, Hamed and Han, Jiawei},
  journal={arXiv preprint arXiv:2503.09516},
  year={2025}
}

@article{elhage2022toy,
  title={Toy models of superposition},
  author={Elhage, Nelson and Hume, Tristan and Olsson, Catherine and Schiefer, Nicholas and Henighan, Tom and Kravec, Shauna and Hatfield-Dodds, Zac and Lasenby, Robert and Drain, Dawn and Chen, Carol and others},
  journal={arXiv preprint arXiv:2209.10652},
  year={2022}
}
